%% file: main.tex
\documentclass[lettersize,journal]{IEEEtran}
\usepackage{amsmath,amsfonts}
\usepackage{algorithmic}
\usepackage{algorithm}
\usepackage{array}
\usepackage[caption=false,font=normalsize,labelfont=sf,textfont=sf]{subfig}
\usepackage{textcomp}
\usepackage{stfloats}
\usepackage{url}
\usepackage{verbatim}
\usepackage{graphicx}
\usepackage{cite}
\usepackage{tikz}
\usepackage{comment}
\usepackage{amssymb} 
\usepackage{color, colortbl}
\usepackage{mathtools}
\usepackage{multirow}
\usepackage{bbding}

\definecolor{gray}{gray}{0.92}

\usepackage[normalem]{ulem}
\usepackage{xcolor}
\usepackage[pagebackref,breaklinks,colorlinks]{hyperref}

\newcommand{\sph}[0]{\mathbb{S}}
\newcommand{\hyp}[0]{\mathbb{H}}

\newcommand{\norm}[1]{\left\lVert #1 \right\rVert}

\newcommand{\R}{\mathbb{R}}

\usepackage{makecell}

\usepackage{arydshln}
\usepackage{caption}
\usepackage{xspace}
\makeatletter
\DeclareRobustCommand\onedot{\futurelet\@let@token\@onedot}
\def\@onedot{\ifx\@let@token.\else.\null\fi\xspace}

\def\eg{\emph{e.g}\onedot} 
\def\ie{\emph{i.e}\onedot} 
 
 \def\vs{\emph{vs}\onedot}

\makeatother

\begin{document}

\title{Curved Geometric Networks for Visual Anomaly Recognition}

\author{Jie Hong, Pengfei Fang$^\dag$, Weihao Li, Junlin Han, Lars Petersson and Mehrtash Harandi
\thanks{J. Hong, P. Fang, and J. Han are with Research School of Electrical, Energy and Material Engineering, the Australian National University, Canberra, ACT 2601, Australia, and are also with Data61-CSIRO, Black Mountain Laboratories, Canberra, ACT 2601, Australia. E-mail: Jie.Hong@anu.edu.au, Pengfei.Fang@anu.edu.au, Junlin.Han@anu.edu.au.}

\thanks{W. Li and L. Petersson are with Data61-CSIRO, Black Mountain Laboratories, Canberra, ACT 2601, Australia. E-mail: Weihao.Li@data61.csiro.au, Lars.Petersson@data61.csiro.au}

\thanks{M. Harandi is with the department of Electrical and Computer Systems Engineering, Monash University, and Data61-CSIRO, Melbourne, Australia. E-mail: mehrtash.harandi@monash.edu}
\thanks{$\dag$: Corresponding author.}
}


\maketitle

\begin{abstract}
Learning a latent embedding to understand the underlying nature of data distribution is often formulated in Euclidean spaces with zero curvature. However, the success of the geometry constraints, posed in the embedding space, indicates that curved spaces might encode more structural information, leading to  better discriminative power and hence richer representations. In this work, we investigate benefits of the curved space for analyzing anomalies or out-of-distribution objects in data. This is achieved by considering embeddings via three geometry constraints, namely, spherical geometry (with positive curvature), hyperbolic geometry (with negative curvature) or mixed geometry (with both positive and negative curvatures). Three geometric constraints can be chosen interchangeably in a unified design given the task at hand. Tailored for the embeddings in the curved space, we also formulate functions to compute the anomaly score. Two types of geometric modules (\ie Geometric-in-One and Geometric-in-Two models) are proposed to plug in the original Euclidean classifier, and anomaly scores are computed from the curved embeddings. We evaluate the resulting designs under a diverse set of visual recognition scenarios, including image detection (multi-class OOD detection and one-class anomaly detection) and segmentation (multi-class anomaly segmentation and one-class anomaly segmentation). The empirical results show the effectiveness of our proposal through the consistent improvement over various scenarios.
\end{abstract}


\input{Intro}
\input{RelatedWork}

\input{Method}
\input{Expt}
\input{Conclusion}

\bibliographystyle{IEEEtran}
\bibliography{bib.bib}

\end{document}

%% file: Intro.tex
\section{Introduction}
In this paper, we aim to leverage the curved geometry for learning embeddings, which in return allows us to analyze and identify anomalies or out-of-distribution (OOD) objects from normal or in-distribution (ID) input data. Non-flat geometry has gained an increasing amount of interest in various machine learning approaches, due to its intriguing properties in encoding the hidden structural information of the data~\cite{nickel2017poincare,tifrea2018poincar,dhingra2018embedding,khrulkov2020hyperbolic,park2021unsupervised,zhang2021learning}.
For example, hyperbolic spaces, featured with a constant negative curvature, are  shown to be rich in encoding the underlying hierarchical structure in the data. Such a property enables hyperbolic spaces to better discriminate input samples~\cite{khrulkov2020hyperbolic}. Spherical spaces with constant positive curvature also show appealing properties along with deep neural networks (DNNs)~\cite{liu2017sphereface,fan2019spherereid}. 

\begin{figure}
\begin{center}
	\includegraphics[width=8.6cm]{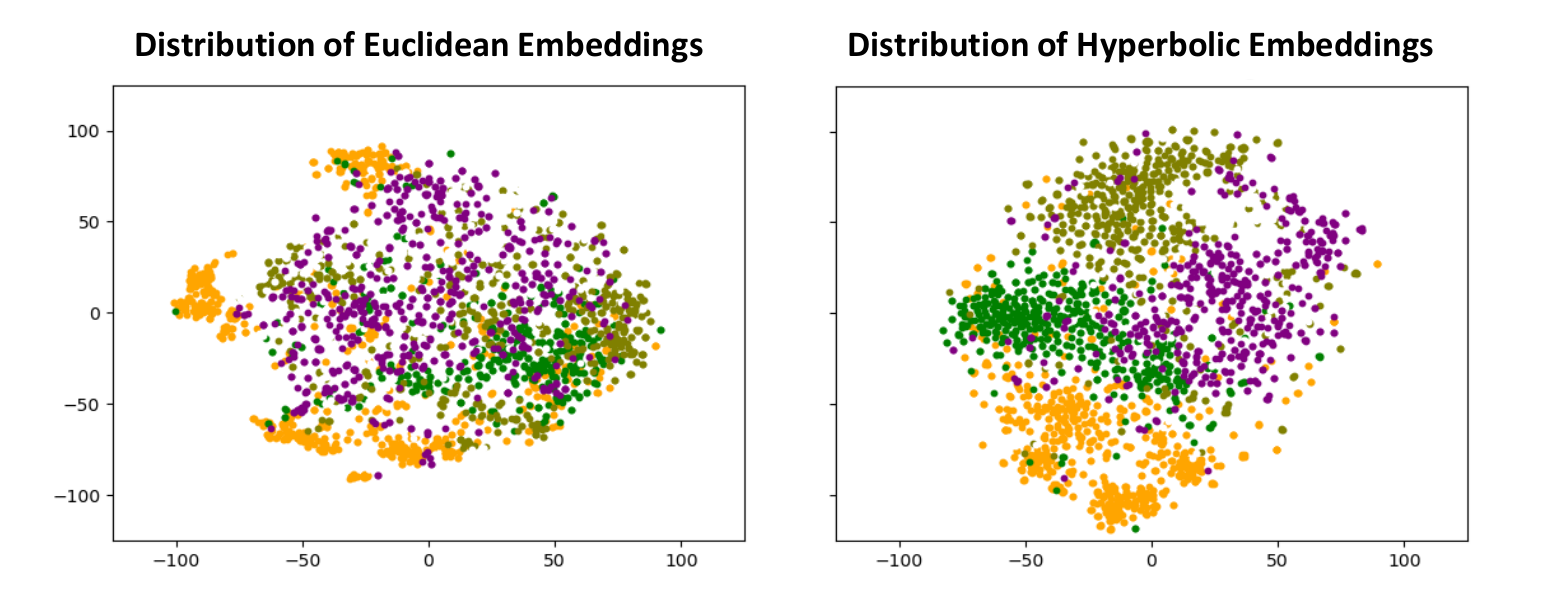}
 	\caption{
 	An illustration of discrimination ability of curved spaces on unknown data. We train two models using Euclidean and hyperbolic geometry on \emph{mini}ImageNet dataset~\cite{ravi2016optimization}, following the few-shot learning protocol. We randomly choose four unknown classes and plot the class embeddings using t-SNE~\cite{van2008visualizing}. 
 	Each unknown class is represented by one color. From this visualization, we can directly observe that within each unknown class, embeddings in hyperbolic space (See the figure on right) show a more compact distribution than embeddings in Euclidean space (See the figure on left). 
 	}\label{fig:intro-1-emb-few}
\end{center}
\end{figure}

Previous studies such as~\cite{liu2020hyperbolic,khrulkov2020hyperbolic,shen2021spherical} show that curved spaces can attain a superior performance gain over the Euclidean space, especially for tasks relying on image embeddings (\eg, zero/few-shot learning or metric learning). 
For example, in~\cite{shen2021spherical}, by employing the similarity metric in spherical embedding spaces, the model enhances its discriminative ability in zero-shot classification for unknown classes. In~\cite{khrulkov2020hyperbolic,fang2021kernel}, hyperbolic spaces are shown to have a better distribution across unknown classes, and therefore improving few-shot learning performance. Also, in~\cite{khrulkov2020hyperbolic}, the model's confidence in the seen samples/unseen samples can also be measured by the geodesic distance in the hyperbolic space.

\begin{figure*}
\begin{center}
	\includegraphics[width=14.0cm]{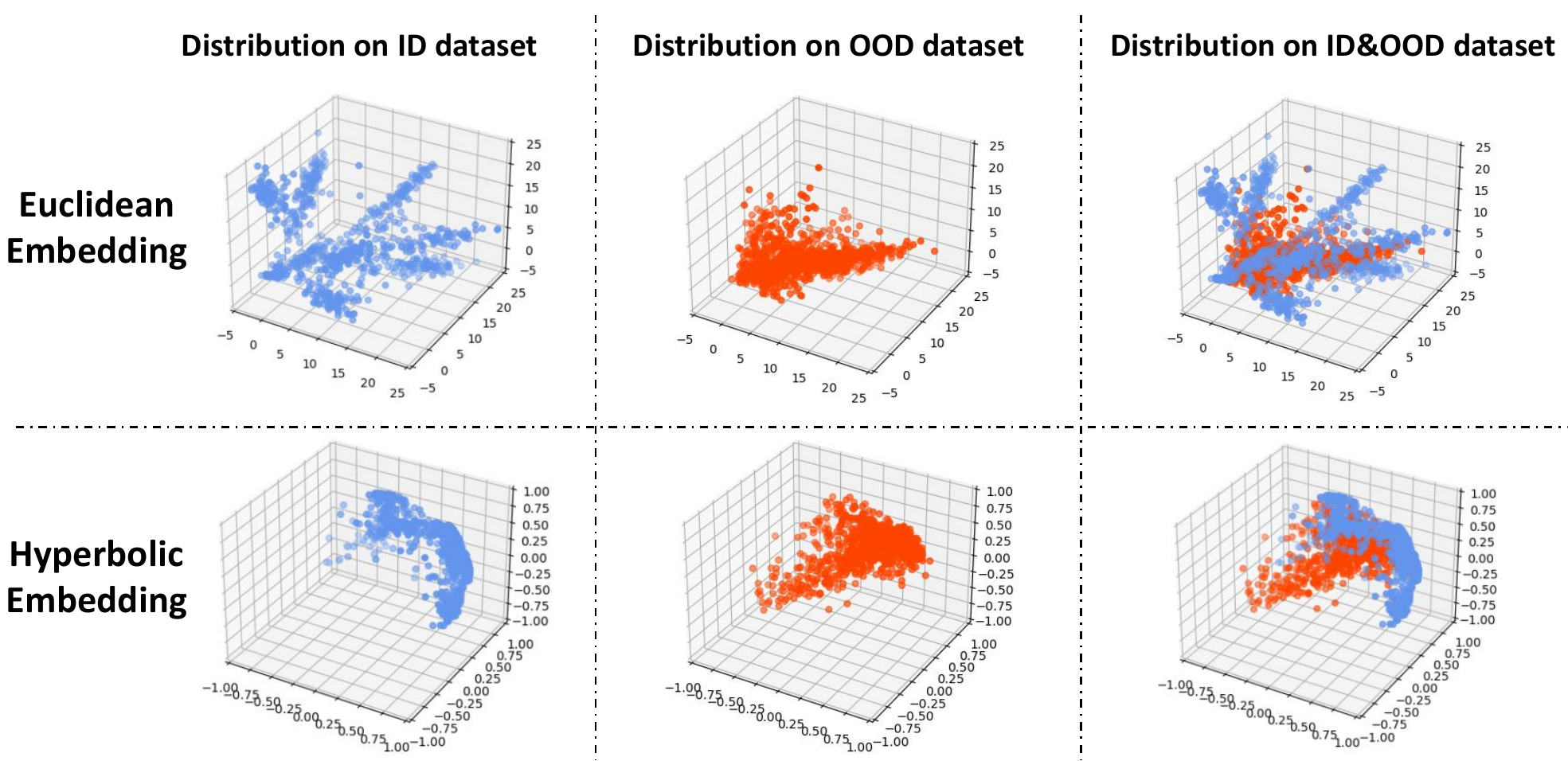}
 	\caption{
 	 An illustration of the ability of curved spaces for distinguishing ID data and OOD data (Image-level OOD detection).
 	 We train two networks using 3-Dimensional Euclidean and hyperbolic embeddings on an ID dataset (CIFAR-10 \cite{krizhevsky2009learning}). We then use the trained models to encode images on both the ID and OOD datasets (the cropped TinyImageNet \cite{deng2009imagenet}). The \textcolor{blue}{`blue'} and \textcolor{red}{`red'} represent the embeddings of ID and OOD inputs, respectively. (1) Left figures: for the ID dataset, both distributions have the basic clustering of intra-classes (ID classes). Euclidean embeddings appear more evenly in the Euclidean space whereas hyperbolic embeddings are mostly distributed near the boundary of the hyperbolic sphere; (2) Middle figures: compared to the ID dataset, the OOD sample points are more densely clustered. Moreover, their points are closer to the origin than ID samples; (3) Right figures: the embedding distributions in the hyperbolic spaces of ID and OOD look more separated than those in the Euclidean space. There is  the less overlap between the \textcolor{blue}{blue} and \textcolor{red}{red} points in the hyperbolic space.
 	}
 	\label{fig:intro-1-emb}
\end{center}
\end{figure*}

The main conjecture is that such spaces can encode complex structured information of the data, thereby improving the discrimination of the embedding. The aforementioned studies also suggest that curved spaces might be able to accommodate compact clusters better, even for unknown data (\ie, small intra-class distance). More specifically,

\begin{itemize}
    \item \textbf{Discrimination ability.} In low-shot learning problems, curved spaces enable the network to learn embedding spaces with smaller within-class variance for unknown samples, leading to a more precise and separated decision boundary among unknown classes. That might lead to a smaller overlap between the embedding distributions of unknown and known classes. See Fig. \ref{fig:intro-1-emb-few} for an illustration. Using the code provided in \cite{snell2017prototypical} and \cite{khrulkov2020hyperbolic}, in this case, we train two models under backbone Conv-4 with 1600-Dimensional Euclidean and hyperbolic embeddings on \emph{mini}ImageNet \cite{ravi2016optimization}. We randomly choose 4 classes from 16 unknown classes to do 2-Dimensional t-SNE visualization \cite{van2008visualizing}. From Fig. \ref{fig:intro-1-emb-few}, we can directly observe that within each unknown class, embeddings in the hyperbolic space (See the right figure) show a more stringent distribution than embeddings in the Euclidean space. The few-shot classification accuracies, 49.42 \% and 54.43\% reported in \cite{snell2017prototypical} and \cite{khrulkov2020hyperbolic}, further indicate that, compared to the Euclidean space, the hyperbolic space is better suited to reduce distances within each unknown class.
    
    \item \textbf{Inhomogeneous property.} In a hyperbolic space, the volume increases exponentially~\cite{chami2021horopca,guo2021co}. Under the assumption that learned embeddings of ambiguous unknown objects tend to be distributed closer to the origin (See the \textcolor{red}{red} samples in Fig. \ref{fig:intro-1-emb}), the inhomogeneous property of the hyperbolic space allows the algorithm to accommodate more embeddings of unknown objects within the expanding space from the origin~\cite{khrulkov2020hyperbolic}. 
\end{itemize}

The above properties serve as inspirations to employ curved geometries to address the problem of visual anomaly recognition. Visual anomaly recognition aims to identify anomalous (or OOD/unknown/unseen) samples from normal (or ID) visual inputs.
Previous studies of anomaly recognition exclusively employ Euclidean spaces to address the problem~\cite{hendrycks2016baseline,vyas2018out,gidaris2018unsupervised,tack2020csi,yi2020patch,Jang2022Collective}.
In order to justify this line of investigation, we first provide a toy example from the task of image-level OOD detection, which attempts to identify OOD and ID images (See Fig. \ref{fig:intro-1-emb}). As shown in Fig. \ref{fig:intro-1-emb}, the embeddings in the hyperbolic space are preferable over the Euclidean space, due to a smaller overlap between the distributions of the ID (\ie, the \textcolor{blue}{blue} points) and the OOD samples (\ie, the \textcolor{red}{red} points).

With the appealing observations shown above, in this paper, we investigate the practice of using \textbf{geometries with fixed non-zero curvature} in \textbf{visual anomaly or OOD recognition} tasks. For the purpose of realizing our idea, a natural solution is to simply replace the existing Euclidean classifier with one based on a curved geometry. This idea is behind the design of our \textit{Geometric-in-One} model. In addition, we also find that the `divergence’ between Euclidean embeddings and curved embeddings can provide a reliable indicator useful for anomaly or OOD identification. To benefit from this interesting observation, we further develop a \textit{Geometric-in-Two} model for anomaly or OOD recognition. Having multiple geometry-aware networks at our disposal, we further present approaches for getting the \textit{anomaly score} to identify abnormal objects. 

Our main objective is to show that the geometry plays an essential role in identifying anomalies. As such, we develop a generic solution and incorporate it into various baselines in our empirical study. The \textbf{contributions} of this work can be summarized as: 

\begin{itemize}
\item[$\bullet$] We propose two types of light-weight curvature-aware geometric networks for visual anomaly or OOD recognition. To the best of our knowledge, this is the first attempt to adopt curved manifolds as embedding spaces to distinguish normal/ID and anomalous/OOD data. Additionally, multiple curved spaces including spherical, hyperbolic, and mixed spaces are studied.

\item[$\bullet$] Extensive experiments on a wide range of visual anomaly or OOD recognition tasks (\eg, multi-class OOD detection, one-class anomaly detection, multi-class anomaly segmentation, and one-class anomaly segmentation) suggest that the proposed technique leads to a substantial performance gain over the Euclidean geometry. 

\end{itemize}

%% file: RelatedWork.tex
\section{Related Work}

\subsection{Visual Anomaly Recognition}
Three main approaches are developed for doing visual anomaly recognition: confidence-, generative-, and self-supervised-based methods. Few works are learning features for those purposes in non-Euclidean spaces.

\textbf{Confidence-based method.} It is well known that the confidence from the softmax in a classifier helps to detect OOD samples from ID samples since ID samples are more likely to have a greater maximum softmax confidence compared to ODD samples \cite{hendrycks2016baseline}. Out-of-Distribution detector for Neural Network (ODIN) \cite{liang2017enhancing} applies temperature scaling to the confidence vector and adds small perturbations to input samples for more accurate OOD detection. Additional confidence-based methods which make use of the confidence have been studied in \cite{devries2018learning,corbiere2019addressing,hendrycks2019scaling,hsu2020generalized,Jang2022Collective}.

\textbf{Generative-based method.} One of the generative-based methods is to synthesize effective training samples to avoid the deep neural networks becoming overconfident in their predictions \cite{lee2017training,sabokrou2018avid,lis2019detecting,xia2020synthesize,Kong_2021_ICCV}. Another choice is to optimize features in the latent space of an encoder-decoder network towards generating a more general distribution \cite{ren2019likelihood,Abati_2019_CVPR,perera2019ocgan,gong2019memorizing,massoli2021mocca} or a more representative attention map \cite{venkataramanan2020attention,liu2020towards,zhou2021memorizing}. 

\textbf{Self-supervised-based method.} Self-supervised learning techniques have been widely employed in anomaly recognition. Ensemble Leave-out Classifier (ELOC) \cite{vyas2018out} trains classifiers in a self-supervised manner by setting a subset of training data as OOD data. One main idea behind the self-learning method is to apply geometric transformations (or augmentations) on the input images and train a multi-class model to discriminate such transformations (or augmentations). Prediction of image rotation is used in Rotation Network (RotNet) \cite{gidaris2018unsupervised}. Jittered patches of an image are classified in Patch-SVDD \cite{yi2020patch} for anomaly localization. Another idea is to use contrastive learning for better visual representations \cite{tack2020csi,zheng2022towards,hong2022goss}. More works using self-supervised learning are presented in \cite{ruff2018deep,golan2018deep,hendrycks2019using,bergman2020classification,tack2020csi,li2021cutpaste,liu2021anomaly}. 

Confidence-based as well as self-supervised-based methods mainly adopt `encoder-classifier' structures while generative-based models are commonly with `encoder-decoder' architectures. The proposed modules in our work are best applied to an `encoder-classifier' rather than an `encoder-decoder' structure.

\subsection{Geometric Learning}
Geometric learning has been studied extensively to encode structured representations. For example, the set has been used to model order-invariant data (\eg, 3D point clouds~\cite{NIPS2017_6931_deepset}, or video data~\cite{Fang_2021_WACV}). Orthogonal constraints,  \ie, subspaces, are often used to encode set data~\cite{Cheraghian_2021_ICCV,Simon_2020_CVPR}, for its potential to be robust against illumination variations, background, etc.  In vectorized representations, spherical or hyperbolic spaces are also very effective for metric learning-related tasks. In the spherical space, the similarity of representations is upper-bounded. Hence, such a space is particularly well-behaved at learning a metric space~\cite{liu2017sphereface,liu2017deep}. As opposed to the spherical space,  tree-like data can be embedded in the hyperbolic space, for its intriguing property to capture the hierarchical structure of the data~\cite{liu2020hyperbolic,khrulkov2020hyperbolic,ma2022adaptive}. To further increase the discrimination power of the learned embeddings in curved spaces, the kernel methods, which implicitly map the geometric representation to a high or even infinite dimensional feature space, are studied for spherical embgedings~\cite{jayasumana2022modelefficient}, or hyperbolic embeddings~\cite{fang2021kernel}. To fully model the structure of the data, mixed-curved spaces are good candidates as embedding spaces~\cite{MixedCuerature_ICLR,MixedVAE_ICLR}.

%% file: Method.tex
\section{Preliminaries and Background}

\begin{figure*}
\begin{center}
	\includegraphics[width=18cm]{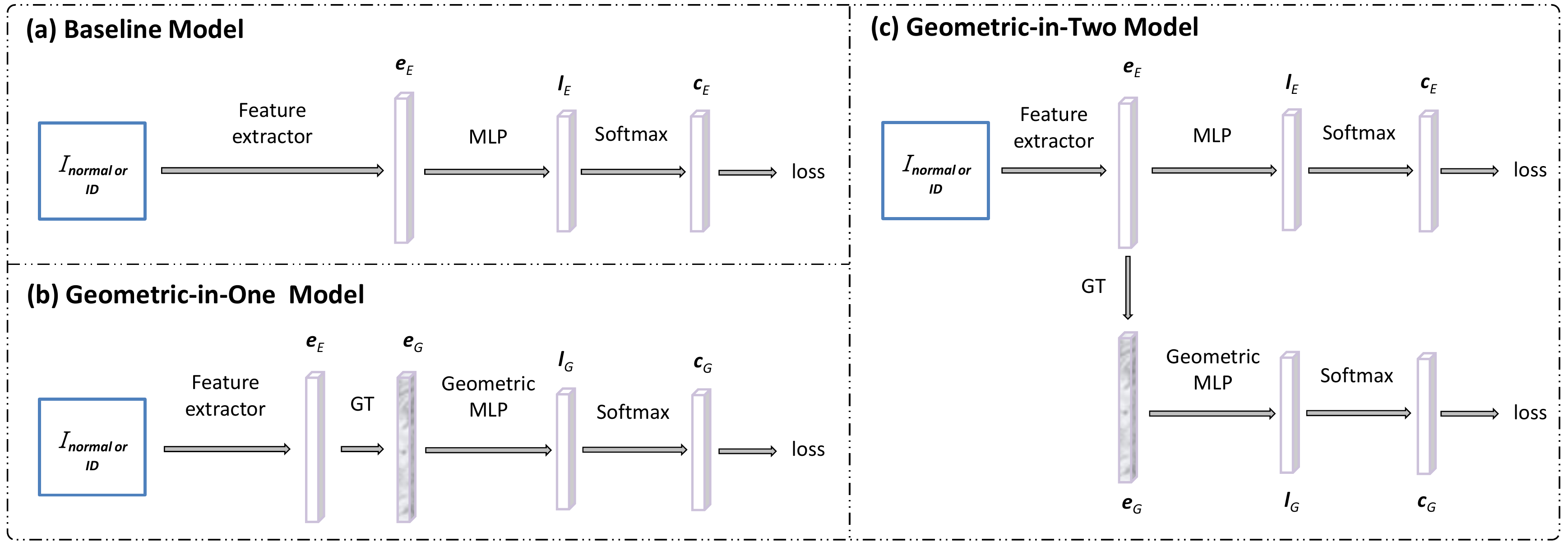}
 	\caption{The general frameworks of the models and their training processes are depicted. We have Geometric-in-One (GiO) and Geometric-in-Two (GiT) models based on the baseline model. In the GiO model, the original Euclidean classifier is replaced with one based on a curved geometry. GiT model adopts the curved classifier as the additional branch. In the training stage, we follow the setting that the training data does not include any anomalous or OOD samples. GT is geometric transformation.
 	}\label{fig:method-1}
\end{center}
\end{figure*}

In this section, we will briefly introduce the preliminary knowledge and background used in this paper.

\subsection{Notation}
We use $\kappa$ to denote the curvature of a manifold. In general, a vectorized representation or an embedding can be embedded in three types of manifolds: the Euclidean space $\mathcal{M}_{E}$, the spherical space $\mathcal{M}_{S}$ and the hyperbolic space $\mathcal{M}_{H}$, corresponding to $\kappa=0$, $\kappa>0$ and $\kappa<0$, respectively. Throughout the paper, we call any space with $\kappa \neq 0$, as a curvature-aware space or a curved space. A mixed-curvature manifold $\mathcal{M}_{M}$ is a product space, consisting of a set of different spaces~\cite{MixedCuerature_ICLR,MixedVAE_ICLR}. In our work, the mixed-curvature manifold is defined as $\mathcal{M}_{M} = \mathcal{M}_{1} \times \mathcal{M}_{2} \times \mathcal{M}_{3} \times ... \times \mathcal{M}_{N}$, in which we mix $N$ different manifolds. 
For example, the mixed-curvature manifold $\mathcal{M}_{M} = \mathcal{M}_{E} \times \mathcal{M}_{S} \times \mathcal{M}_{H}$ includes an Euclidean space, a spherical space and a hyperbolic space.

\subsection{Spherical Geometry}
The $n$-sphere with curvature $\kappa > 0$ 
is defined as 
\begin{align}
\sph^{n-1}_\kappa = \{\textbf{x} \in \R^{n} : \norm{\textbf{x}}^2 = 1/\kappa\}\;.
\label{eqn:projecvtion_sphere}
\end{align}
The mapping $\Gamma_\sph:\mathbb{R}^n \to \sph_{\kappa}^{n-1}$ projects an embedding $\textbf{x} \in \mathbb{R}^n$ generated by a feature extractor to $n$-sphere as:
\begin{align}
    \textbf{x}_{S} = \Gamma_\sph(\textbf{x}) = \frac{\textbf{x}}{\sqrt{\kappa}\|\textbf{x}\|}.
    \label{eqn:ST}
\end{align}

In practice, the angular mapping in the $n$-sphere, analogous to the linear mapping in the Euclidean space, can be realized by a fully-connected (FC) layer. In such a case, let $\textbf{W} = \big[\textbf{w}_1, \textbf{w}_2, \cdots, \textbf{w}_C\big]$ be a matrix storing $C$ class prototypes with $\textbf{w}_i \in \sph_{\kappa}^{n-1}$. The angular mapping is simply the inner product between $\textbf{x} \in \sph_{\kappa}^{n-1}$ and columns of $\textbf{W}$. We note that for $\textbf{x}, \textbf{w}_j \in \sph_{\kappa}^{n-1}$, the angular mapping for the $j$th class $l_{\sph}(\textbf{x}, \textbf{w}_j) = \langle \textbf{x}, \textbf{w}_j\rangle$ is indeed related to the geodesic distance on $\sph_{\kappa}^{n-1}$, hence one can understand the angular mapping as a form of distance-based method. We use the notation ${l}_{\sph}(\textbf{x}, \textbf{W})$ to show a vector obtained by applying the columns of $\textbf{W}$ to $\textbf{x} \in \sph_\kappa^{n-1}$.

For the spherical network, we compute the angular loss $\ell_{\sph}$ based on $B$ samples in one batch:
\begin{equation}
\begin{aligned}
\ell_{\sph} = -\frac{1}{B}\sum_{i}\mathrm{log}\frac{\exp^{l_{\sph,y_i}}}{\sum_{j} \exp^{l_{\sph,j}}}
\end{aligned}\label{SLoss}
\end{equation}
where $l_{\sph,j}$ is the $j$th element in $l_{\sph}(\textbf{x}, \textbf{W})$ under the $i$th input sample and $\textbf{x} \in \sph_{\kappa}^{n-1}$. Accordingly, $l_{\sph,{y}_i}$ is the the $y_i$th element in $l_{\sph}(\textbf{x}, \textbf{W})$ and $y_i$ indicates the label class to the $i$th input sample.

\subsection{Hyperbolic Geometry}
In contrast to the $n$-sphere $\sph_{\kappa}^{n-1}$, the hyperbolic space is a curved space with a constant negative curvature (i.e., $\kappa < 0$). In this paper, we employ the Poincar\'e ball \cite{nickel2017poincare,khrulkov2020hyperbolic} to model and work with the hyperbolic space. The $n$-dimensional Poincar\'e ball, with curvature $\kappa$, is defined by the manifold 

\[
\hyp_{\kappa}^n = \{ \textbf{x} \in \R^n: \norm{\textbf{x}} < -1/\kappa \}\;\footnote{In this case, the Riemannian metric is defined as $g_{\kappa}^{\hyp}(\textbf{x}) = \lambda^2_{\kappa}(\textbf{x})\cdot g^{E}$, where $\lambda_{\kappa}(\textbf{x}) = \frac{1}{1+{\kappa}\norm{\textbf{x}}^2}$, and $g^E$ is the Euclidean metric.}.
\]
To embed $\textbf{x} \in \R^n$, obtained by a feature extractor to the Poincar\'e ball, we use the following transformation:
\begin{equation}    \label{eqn: clip norm implement}
\textbf{x}_{H}= \Gamma_\hyp(\textbf{x}) =
\begin{cases}
    \textbf{x}   &  \text{if}~~\|\textbf{x} \| \leq \frac{1}{|\kappa|} \\
    \frac{1-\xi}{|\kappa|} \frac{\textbf{x}}{ \norm{\textbf{x}}}
        &  \text{else},
\end{cases}                
\end{equation}
where $\xi > 0$ is a small value to ensure numerical stability. To enable the vector operations in the Poincar\'e ball, we make use of  the M\"obius addition for $\textbf{x},\textbf{y} \in \hyp^{n}_\kappa$ as:
\begin{equation}
\begin{aligned}
\textbf{x} \oplus_{\kappa} \textbf{y} = \frac{(1 + 2|\kappa| \langle \textbf{x}, \textbf{y}\rangle + |\kappa| \|\textbf{y}\|^2)\textbf{x} + (1-|\kappa| \|\textbf{x} \|^2)\textbf{y}}{1 +2|\kappa| \langle \textbf{x}, \textbf{y} \rangle + |\kappa|^2 \|\textbf{x}\|^2\|\textbf{y}\|^2},
\end{aligned}
\end{equation}
where $\langle , \rangle$ is the inner product. The geodesic distance between $\textbf{x},\textbf{y} \in \hyp^{n}_\kappa$ is defined as:
\begin{equation}
d_{Geo}(\textbf{x}, \textbf{y}) = \frac{2}{\sqrt{|\kappa|}}\mathrm{tanh}^{-1}(\sqrt{|\kappa|}\|-\textbf{x} \oplus_{\kappa} \textbf{y}\|).
\label{eq:geodis}
\end{equation}

One can also generalize the hyperbolic linear operation, parameterized by $\textbf{W}$ (\eg, the hyperbolic linear layer), in the Poincar\'e ball \cite{khrulkov2020hyperbolic}:
\begin{equation}
\textbf{W}_{\oplus_{\kappa}}(\textbf{x}) \coloneqq \frac{1}{\sqrt{|\kappa|}} \mathrm{tanh}\big(\frac{\|\textbf{W}\textbf{x}\|}{\|\textbf{x}\|}\mathrm{tanh}^{-1}(\sqrt{|\kappa|}\|\textbf{x}\|)\big) \frac{\textbf{W}\textbf{x}}{\|\textbf{W}\textbf{x}\|}.
\end{equation}

The proposed network contains the multi-class classification layer. We employ the generalization of multi-class logistic regression (MLR) to the hyperbolic spaces~\cite{khrulkov2020hyperbolic}. Following the work in~\cite{khrulkov2020hyperbolic}, the formulation of the hyperbolic MLR for $C$ classes is given by:
\begin{equation}
\begin{aligned}
&l_{H}(y = j | \textbf{x}) \propto 
\\
&\exp\Big( \frac{\lambda^{\kappa} (\textbf{x})\|\textbf{W}_{j}\|}{\sqrt{|\kappa|}} \mathrm{sinh}^{-1} \big( \frac{2\sqrt{|\kappa|} \langle -\textbf{p}_{j} \oplus_{\kappa} \textbf{x}, \textbf{W}_{j}\rangle}{(1-|\kappa| \cdot \|-\textbf{p}_{j} \oplus_{\kappa} \textbf{x}\|^2)\|\textbf{W}_{j}\|} \big)\Big)
\end{aligned}
\label{HLoss}
\end{equation}
where $j \in \{1, 2, ..., C \}$. Here,  $\textbf{x}\in \hyp^{n}_\kappa$ is an embedding in the hyperbolic space, and $\textbf{p}_{j} \in \hyp_{\kappa}^n$, $\textbf{W}_{j} \in T_{\textbf{p}_{\kappa}}\hyp_{\kappa}^n \backslash \{\textbf{0}\}$ are learnable weights.

\begin{figure*}
\begin{center}
	\includegraphics[width=14.0cm]{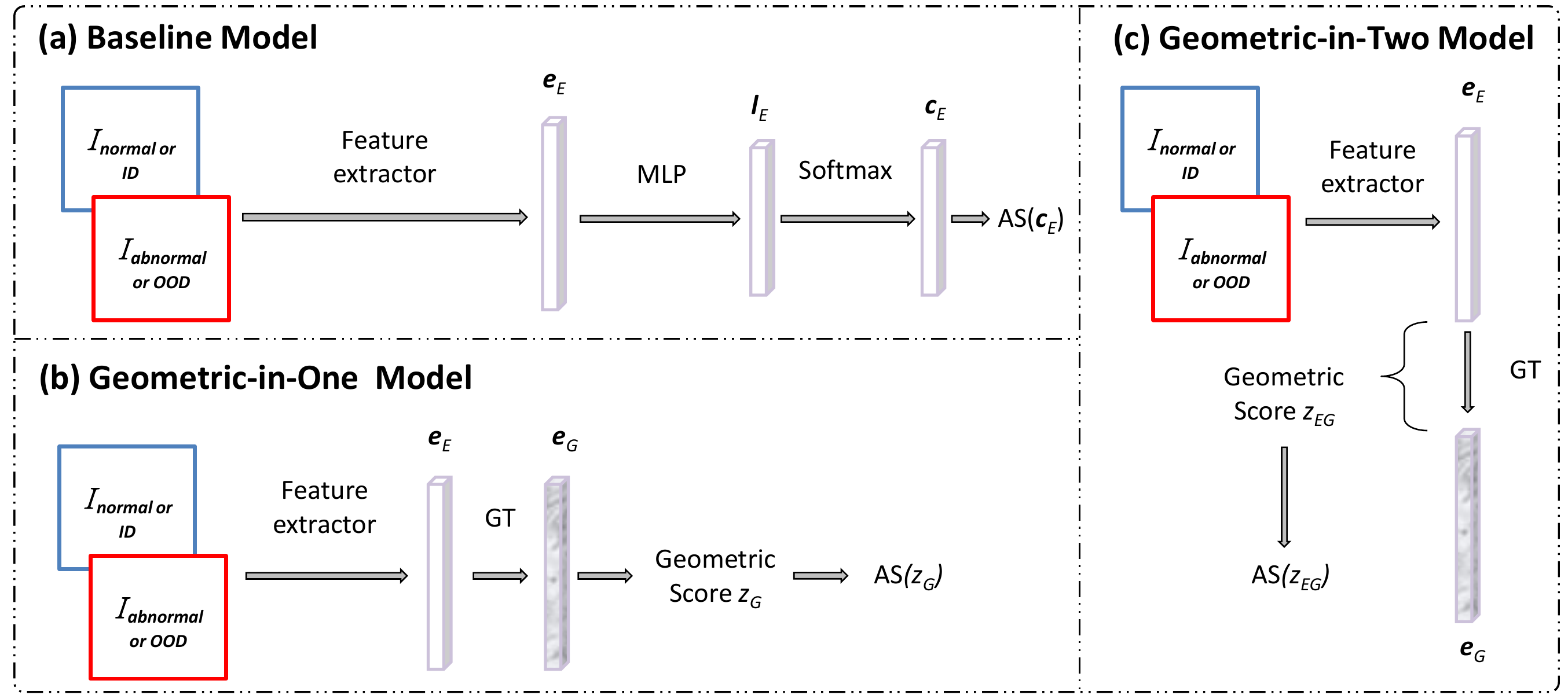}
 	\caption{The general frameworks of the models and their evaluation processes are depicted. In the evaluation stage, the model would processes both normal (or ID) and anomalous (or OOD) inputs. The geometric score $z_{G}$ is extracted from the curved embedding for anomaly score computation in GiO model. For GiT model, the geometric score $z_{EG}$ is obtained via Euclidean and curved embeddings. During the evaluation process, GiO and GiT models use the less parameters than the baseline.
 	}\label{fig:method-2}
\end{center}
\end{figure*}

\section{Approach}
Visual anomaly recognition aims to identify abnormal (or OOD) samples from normal (or ID) samples. During the training process, as illustrated in Fig. \ref{fig:method-1}, only normal or ID data can be accessed. For the evaluation stage, as shown in Fig. \ref{fig:method-2}, both normal (or ID) and anomalous (or OOD) inputs to be recognized, exist.

The pipeline of the baseline is illustrated in Fig. \ref{fig:method-1} (a) for the training phase and \ref{fig:method-2} (a) for the inference phase. Specifically, a feature extractor first maps the input to a feature embedding $\textbf{e}_E$, lying in a Euclidean space. The following MLP layer and $\mathrm{Softmax}$ function further predict the probability belonging to each normal (or ID) class, denoted by $\textbf{c}_E$ (See Fig. \ref{fig:method-1} (a)). In the evaluation process, to identify whether the input image or pixel is an outlier (a.k.a., anomalous or OOD data), one needs to define the anomaly score $\text{AS} \in [0, 1]$. In this vanilla model (See Fig. \ref{fig:method-2} (a)), we follow the common practice in \cite{hendrycks2016baseline} to define the anomaly score by leveraging the predication $\textbf{c}_E$ as $\text{AS} = 1-\mathrm{max}(\textbf{c}_E)$.

A curvature-aware geometric model indicates a model where its classifier operates in a curved space $\mathcal{M}_{G}$ and we term its classifier geometric classifier. Two curvature-aware geometric models are presented in this section: Geometric-in-One (GiO) and Geometric-in-Two (GiT), as shown in Fig. \ref{fig:method-1} (b) and (c). Compared to the baseline model, the curvature-aware geometric model has two geometric layers, namely, geometric transformation 

and geometric MLP. The geometric transformation is to transform the Euclidean embedding $\textbf{e}_{E}$ computed by a feature extractor to the geometric embedding $\textbf{e}_{G}$. See Eq. \eqref{eqn:ST} for the spherical geometry $\mathcal{M}_{S}$ and Eq. \eqref{eqn: clip norm implement} for the hyperbolic geometry $\mathcal{M}_{H}$. The geometric MLP is a generalization of MLP in $\mathcal{M}_{G}$ (\eg, the angular linear layers in $\mathcal{M}_{S}$ or the hyperbolic linear layers in $\mathcal{M}_{H}$). As shown in Fig. \ref{fig:method-1}, we learn geometric classifiers where embeddings extracted from a feature extractor are manipulated in the curved spaces. 
As explained in Fig. \ref{fig:method-2} (b) and (c), in contrast to the baseline model, during the inference phase, the anomaly score in our geometric model is defined as $\text{AS} = 1-\mathrm{tanh}(z)$. The geometric score $z$ is a framework-dependent value whose details will be introduced below.

\subsection{Geometric-in-One Model}
We first introduce a single-branch framework, \ie,  the Geometric-in-One (GiO) model. The GiO model (See Fig. \ref{fig:method-1} (b)) is a natural modification that replaces the Euclidean embeddings with curved-geometry embeddings. It provides a straightforward comparison to justify the advantages of using curved geometry for anomaly recognition. Specifically, as shown in Fig. \ref{fig:method-1} (b), the image or pixel embeddings can be obtained by applying the geometric transformation function (\ie, Eq. \eqref{eqn:ST} for $\mathcal{M}_{S}$ and Eq. \eqref{eqn: clip norm implement} for $\mathcal{M}_{H}$) to the feature vectors encoded by the feature extractor. Then a following geometric classifier, realized by geometric MLP, is further used to predict the class of its input. For example, the Hyperbolic-in-One model refers to a network with a geometric classifier in the hyperbolic space. Compared to the baseline model (See Fig. \ref{fig:method-1} (a)), GiO model only modifies the embedding layer and classifier without bringing extra parameters, thereby being a cheap yet flexible solution.  

In the GiO model, the geometric score $z_G \in \{z_{S}, z_{H}, z_{M} \}$ can be obtained from embedding $\textbf{e}_{G}$. In our work, the geometric score from the spherical manifold $\mathcal{M}_{S}$ is defined as
\begin{align}\label{eq:gs_s}
z_{S}(\textbf{e}_{S}) = \max({l}_{\sph}(\textbf{e}_{S}, \textbf{W})),
\end{align}
where $\textbf{e}_{S} \in \sph_{\kappa}^{n-1}$. Experiments in \cite{liu2017sphereface,massoli2021mocca} verified that $z_{S}$ from $\mathcal{M}_{S}$ is suitable for visual tasks under the open-set protocol. Hence, we expect $z_{S}$ could help in anomaly or OOD recognition. For the hyperbolic space $\mathcal{M}_{H}$, the geometric score is defined as
\begin{align}\label{eq:gs_h}
z_{H}(\textbf{e}_{H}) = d_{Geo}(\textbf{e}_{H}, \textbf{0}_{H}),
\end{align}
where $d_{Geo}(\textbf{e}_{H}, \textbf{0}_{H})$ is also known as the geodesic distance (See Fig. \ref{eq:geodis}) between the point $\textbf{e}_{H}$ and the origin $\textbf{0}_{H}$, for $\textbf{e}_{H}, \textbf{0}_{H} \in \hyp^{n}_\kappa$. Experiment of image-level OOD detection in Fig. \ref{fig:intro-1-emb} shows the property of $\textbf{e}_{H}$, the hyperbolic embedding points of OOD samples are closer to the origin. A similar observation is also made in~\cite{khrulkov2020hyperbolic}. In the mixed-curvature manifold $\mathcal{M}_{M}$, the geometric score $z_{M}$ is formulated as:
\begin{align} 
  z_{M}^2 = \sum_{i}^{N} z_{M,i}^2,
  \label{eqn:dM}
\end{align}
where there are $N$ spaces in $\mathcal{M}_{M}$ and $z_{M,i}$ is the geometric score from the $i$th component space. For instance, in $\mathcal{M}_{M} = \mathcal{M}_{S} \times \mathcal{M}_{H}$, $z_{M}^2 = z_{S}^2 + z_{H}^2$. The GiO model with $\mathcal{M}_{M} = \mathcal{M}_{S} \times \mathcal{M}_{H}$ is actually a two-branch architecture. We can even incorporate the Euclidean space into the Mixed-in-One model. In $\mathcal{M}_{M} = \mathcal{M}_{E} \times \mathcal{M}_{S} \times \mathcal{M}_{H}$, we can have $z_{M}^2 = z_{E}^2 + z_{S}^2 + z_{H}^2$ where the geometric score from Euclidean space is $z_{E} = \mathrm{max}(\textbf{c}_E)$ \cite{hendrycks2016baseline}. Tab.~\ref{table:GIO} lists the proposed geometric networks, in conjunction with the geometry score, in the GiO model.

Having the geometry score $z_G$ at our disposal, the anomaly score is defined as $\text{AS} = 1-\tanh(z_G)$. A higher value of $\text{AS}$ indicates a higher probability that the input is coming from the anomalous distribution. The purpose of $\tanh(\cdot)$ is to normalize $z_G$ to a value which is in the range $0$ and $1$ (See Fig. \ref{fig:method-2} (b)).

\begin{table}[!ht]
\caption{\textbf{The proposed geometric networks} in the Geometric-in-One (GiO) Model and the Geometric-in-Two (GiT) Model. ``Abb.'', ``Geo. Components'' and ``Geo. Score'' indicate ``abbrevation'', ``geometry componets'' and ``geometry score'' respectively.}
\label{table:GIO}
\centering
\scalebox{1}{
\begin{tabular}{l c c c}
\hline
Method & Abb. & Geo. Components & Geo. Score\\
\hline
\multicolumn{4}{c}{Geometric-in-One (GiO) Model}\\
\hline
Spherical-in-One & SiO & $\mathcal{M}_S$ & Eq.~\eqref{eq:gs_s}\\
Hyperbolic-in-One & HiO & $\mathcal{M}_H$ & Eq.~\eqref{eq:gs_h}\\
Mixed-in-One & MiO & Multiple $\mathcal{M}$ & Eq.~\eqref{eqn:dM}\\
\hline
\multicolumn{4}{c}{Geometric-in-Two (GiT) Model}\\
\hline
Spherical-in-Two & SiT & $\mathcal{M}_S$ & Eq.~\eqref{eqn:dEG}\\
Hyperbolic-in-Two & HiT & $\mathcal{M}_H$ & Eq.~\eqref{eqn:dEG}\\
Mixed-in-Two & MiT & Multiple $\mathcal{M}$ & Eq.~\eqref{eqn:dEM}\\
\hline
\end{tabular}
}
\vspace{-6pt}
\end{table}

\begin{table*}[!ht]
\caption{\textbf{Visual anomaly tasks} where we evaluate the proposed geometric models.
}
\label{table:tasks}
\centering
\scalebox{1}{
\begin{tabular}{l c c c c}
\hline
\textbf{Task} & \textbf{Recognition Space} & \textbf{Num. normal or ID Class} & \textbf{Num. abnormal or OOD Class} \\
\hline
\textbf{Multi-Class OOD Detection}        & Image & Multiple & Multiple \\
\textbf{One-Class Anomaly Detection}      & Image & One & Multiple\\
\textbf{Multi-Class Anomaly Segmentation} & Pixel & Multiple & Multiple\\
\textbf{One-Class Anomaly Segmentation}   & Pixel & One & Multiple\\
\hline
\end{tabular}
}
\end{table*}

\begin{table*}
\begin{center}
\caption{\textbf{Multi-class OOD detection} on CIFAR-10/CIFAR-100 \cite{krizhevsky2009learning} with DenseNet/WRN-28-10 backbones. Mixed-geometry embedding in the +MiO/+MiT model includes spherical embedding and hyperbolic embedding. We provide averaging results over 5 multiple OOD datasets: TINc, TINr, LSUNc, LSUNr and iSUN. ``*" indicates results are obtained via self-implemented network.} 
\label{table:cifar10_100}

\resizebox{0.9\textwidth}{!}{
\begin{tabular}{l|c|c|c|c}
\hline
\makecell[c]{\textbf{Backbone} \\ \textbf{(on ID dataset)}} &\textbf{Method} &\textbf{FPR (95\% TPR) $\downarrow$} &\textbf{Detection Error $\downarrow$} &\textbf{AUROC $\uparrow$} 
\\ \hline

\multirow{3}*{\makecell[c]{\textbf{Dense-BC} \\ \textbf{(on CIFAR-10)}}} &\textbf{Hendrycks$\&$Gimpel}$^*$ \cite{hendrycks2016baseline}  
                                           &36.9 &10.9 &94.8 \\
&\cellcolor{gray}\textbf{+SiO/+HiO/+MiO}   &\cellcolor{gray}37.8/6.3/8.3   &\cellcolor{gray}11.0/5.4/6.2  &\cellcolor{gray}94.7/98.7/98.5 \\
&\cellcolor{gray}\textbf{+SiT/+HiT/+MiT}   &\cellcolor{gray}10.5/20.8/11.2 &\cellcolor{gray}6.7/10.0/7.0  &\cellcolor{gray}98.1/96.0/98.0
\\ \hline

\multirow{3}*{\makecell[c]{\textbf{Dense-BC} \\ \textbf{(on CIFAR-100)}}} &\textbf{Hendrycks$\&$Gimpel}$^*$ \cite{hendrycks2016baseline}  
                           &70.7 &26.3 &81.5 \\
&\cellcolor{gray}\textbf{+SiO/+HiO/+MiO}   &\cellcolor{gray}69.4/43.0/52.4 &\cellcolor{gray}26.1/17.1/21.5 &\cellcolor{gray}81.4/90.7/86.8 \\
&\cellcolor{gray}\textbf{+SiT/+HiT/+MiT}   &\cellcolor{gray}52.6/55.9/66.4 &\cellcolor{gray}20.3/21.8/25.7 &\cellcolor{gray}87.5/85.9/81.7
\\ \hline

\multirow{3}*{\makecell[c]{\textbf{WRN-28-10} \\ \textbf{(on CIFAR-10)}}} &\textbf{Hendrycks$\&$Gimpel}$^*$ \cite{hendrycks2016baseline}  
                           &41.9 &12.5 &93.5 \\
&\cellcolor{gray}\textbf{+SiO/+HiO/+MiO}   &\cellcolor{gray}43.7/16.4/17.9 &\cellcolor{gray}12.7/8.5/9.3 &\cellcolor{gray}93.4/97.1/96.9 \\
&\cellcolor{gray}\textbf{+SiT/+HiT/+MiT}   &\cellcolor{gray}15.5/28.7/20.1 &\cellcolor{gray}8.1/12.5/9.8  &\cellcolor{gray}97.1/94.1/96.3
\\ \hline

\multirow{3}*{\makecell[c]{\textbf{WRN-28-10} \\ \textbf{(on CIFAR-100)}}} &\textbf{Hendrycks$\&$Gimpel}$^*$ \cite{hendrycks2016baseline}  
                           &70.8 &26.4 &81.5 \\
&\cellcolor{gray}\textbf{+SiO/+HiO/+MiO}   &\cellcolor{gray}69.9/51.1/49.2 &\cellcolor{gray}26.3/20.3/19.2 &\cellcolor{gray}81.5/88.1/88.8 \\
&\cellcolor{gray}\textbf{+SiT/+HiT/+MiT}   &\cellcolor{gray}49.3/56.4/57.1 &\cellcolor{gray}20.0/21.7/21.8 &\cellcolor{gray}88.2/85.7/86.4
\\ \hline

\end{tabular}
}
\end{center}
\end{table*}

\begin{table*}
\begin{center}
\caption{\textbf{Multi-class OOD detection} on CIFAR-10/CIFAR-100 \cite{krizhevsky2009learning} with DenseNet/WRN-28-10 backbones. We provide averaging results over 5 multiple OOD datasets: TINc, TINr, LSUNc, LSUNr and iSUN. ``*" indicates results are obtained via self-implemented networks. 
}
\label{table:cifar10_100_odin}

\resizebox{0.75\textwidth}{!}{
\begin{tabular}{l|c|c|c|c}
\hline
\makecell[c]{\textbf{Backbone} \\ \textbf{(on ID dataset)}} &\textbf{Method} &\textbf{FPR (95\% TPR) $\downarrow$} &\textbf{Detection Error $\downarrow$} &\textbf{AUROC $\uparrow$} 
\\ \hline

\multirow{2}*{\makecell[c]{\textbf{Dense-BC} \\ \textbf{(on CIFAR-10)}}} &\textbf{ODIN}$^*$ \cite{liang2017enhancing}  
                                           &11.2 &6.8 &98.0 \\
&\cellcolor{gray}\textbf{+HiO}   &\cellcolor{gray}6.0   &\cellcolor{gray}5.2  &\cellcolor{gray}98.8 \\ \hline

\multirow{2}*{\makecell[c]{\textbf{Dense-BC} \\ \textbf{(on CIFAR-100)}}} &\textbf{ODIN}$^*$ \cite{liang2017enhancing}  
                           &48.0 &17.3 &90.3 \\
&\cellcolor{gray}\textbf{+HiO}   &\cellcolor{gray}40.9 &\cellcolor{gray}16.5 &\cellcolor{gray}91.1 \\ \hline

\multirow{2}*{\makecell[c]{\textbf{WRN-28-10} \\ \textbf{(on CIFAR-10)}}} &\textbf{ODIN}$^*$ \cite{liang2017enhancing}  
                           &22.9 &10.1 &95.8 \\
&\cellcolor{gray}\textbf{+HiO}   &\cellcolor{gray}15.5 &\cellcolor{gray}8.3 &\cellcolor{gray}97.3 \\ \hline

\multirow{2}*{\makecell[c]{\textbf{WRN-28-10} \\ \textbf{(on CIFAR-100)}}} &\textbf{ODIN}$^*$ \cite{liang2017enhancing}  
                           &52.2 &19.2 &88.8 \\
&\cellcolor{gray}\textbf{+HiO}   &\cellcolor{gray}49.5 &\cellcolor{gray}20.0 &\cellcolor{gray}88.3 \\ \hline

\end{tabular}
} 
\end{center}
\end{table*}

\subsection{Geometric-in-Two Model}
Several recent works \cite{yu2019unsupervised,bergmann2020uninformed,Salehi_2021_CVPR,Wang_2021_CVPR} develop the dual-branch architecture and exploit the discrepancy between features in separate classifiers for anomaly or OOD recognition. Motivated by this fact, we further introduce our second framework, termed Geometric-in-Two (GiT), where a Euclidean classifier and a geometric classifier are integrated after the feature extractor (See Fig. \ref{fig:method-1} (c)). In the GiT model, in parallel with a branch of the Euclidean classifier, the other branch learns the feature embedding in the curved space, and a following geometric MLP is used as a class predictor. The embedding in the curved space $\mathcal{M}_{G}$ is achieved by transforming $\textbf{e}_{E}$ to $\textbf{e}_{G}$ via a GT function, as shown in Fig. \ref{fig:method-1} (c). In such a pipeline, the geometry-aware score $z_{EG}$ is defined as the discrepancy between distributions of $\textbf{e}_{E}$ and $\textbf{e}_{G}$, measured via the Kullback-Leibler (KL) divergence, as follows:
\begin{align}\label{eqn:dEG}
z_{EG} = \sum_{i} p_{E,i} \mathrm{log}\frac{p_{E,i}}{p_{G,i}},
\end{align}
where $p_{E,i}$ and $p_{G,i}$ are the $i$th element in $\textbf{p}_E$ and $\textbf{p}_G$. Here, $\textbf{p}_E = \mathrm{softmax}(\textbf{e}_E)$ and $\textbf{p}_G = \mathrm{softmax}(\textbf{e}_G)$. The geometric score $z_{EG}$ is essentially the distribution discrepancy between the learned embedding $\textbf{e}_E$ from $\mathcal{M}_{E}$ and $\textbf{e}_G$ from $\mathcal{M}_{G}$. We have three types of GiT models: Spherical-in-Two, Hyperbolic-in-Two and Mixed-in-Two models, thereby $z_{EG} \in \{ z_{ES}, z_{EH}, z_{EM} \}$ where $z_{ES}$, $z_{EH}$ and $z_{EM}$ are from three models, respectively. The value $z_{ES}$ and $z_{EH}$ can be easily calculated by Eq.~\eqref{eqn:dEG}. Inspired by Eq. \eqref{eqn:dM}, we define $z_{EM}$ in $\mathcal{M}_{M}$ as follows:
\begin{align}\label{eqn:dEM}
z_{EM}^2 = \sum_{i}^{N} z_{EM,i}^2,
\end{align}
where $N$ indicates the number of component spaces in $\mathcal{M}_{M}$. For example, when $\mathcal{M}_{M} = \mathcal{M}_{S} \times \mathcal{M}_{H}$, the score metric can be obtained by $z_{EM}^2 = z_{ES}^2 + z_{EH}^2$. The GiT model with $\mathcal{M}_{M} = \mathcal{M}_{S} \times \mathcal{M}_{H}$ is actually a three-branch architecture.

Our experiments show that $z_{EG}$ is able to provide reliable discrimination information for anomaly identification. We find that, similar to our GiO models, the anomaly score of GiT models should be $\text{AS} = 1-\mathrm{tanh}(z_{EG})$ (See Fig. \ref{fig:method-2} (c)). Tab.~\ref{table:GIO} lists the networks and the geometry score in the GiT model.

\subsection{Model Training}
In the baseline model, as shown in Fig. \ref{fig:method-1} (a), the Euclidean classifier is optimized by a standard cross-entropy loss function, as $\ell = \ell_{E}(\textbf{c}_E)$. Similarly, we optimize the GiO model using the confidence vector $\textbf{c}_G$, predicted in its geometric classifier with its own specific loss $\ell = \ell_{G}(\textbf{c}_G)$ (See Fig. \ref{fig:method-1} (b)). The loss functions for the spherical and hyperbolic geometric networks are described in Eqs. \eqref{SLoss} and \eqref{HLoss}.

The GiT model, as shown in Fig. \ref{fig:method-1} (c), is trained in a multi-task learning manner by optimizing a Euclidean classifier and a geometric classifier, as $\ell = \ell_E(\textbf{c}_E) + \ell_G(\textbf{c}_G)$. To be more specific, a shared feature extractor encodes the input image in a Euclidean space $\mathcal{M}_{E}$ and the curved spaces $\mathcal{M}_{G}$. Then, the following Euclidean classifier and geometric classifier are optimized separately. In contrast to the well-studied student-teacher models~\cite{bergmann2020uninformed, Salehi_2021_CVPR}, which aim to transfer the knowledge from the teacher model to the student model, our GiT model learns sub-branches guided by its own spaces and objective functions (See Eqs. \eqref{SLoss} and \eqref{HLoss}).

%% file: Expt.tex
\section{Experiments}
\begin{table*}
\begin{center}
\caption{\textbf{Multi-class OOD detection} on CIFAR-100 \cite{krizhevsky2009learning} with DenseNet \cite{huang2017densely} and WRN-28-10 \cite{zagoruyko2016wide} backbones. Mixed-geometry embedding in the +MiT model combines a spherical embedding and hyperbolic embedding. We provide averaging results over 4 multiple OOD datasets: TINc, TINr, LSUNc, and LSUNr.}
\label{table:cifar10_100_eloc}

\resizebox{0.85\textwidth}{!}{
\begin{tabular}{l|c|c|c|c}
\hline
\makecell[c]{\textbf{Backbone} \\ \textbf{(on ID dataset)}} &\textbf{Method} &\textbf{FPR (95\% TPR) $\downarrow$} &\textbf{Detection Error $\downarrow$} &\textbf{AUROC $\uparrow$} 
\\ \hline

\multirow{2}*{\makecell[c]{\textbf{Dense-BC} \\ \textbf{(on CIFAR-100)}}} &\textbf{ELOC} \cite{vyas2018out}  
                                           &14.93 &8.37 &97.28 \\
&\cellcolor{gray}\textbf{+SiT/+HiT/+MiT}   &\cellcolor{gray}17.84/10.45/12.84 &\cellcolor{gray}9.20/6.78/7.75 &\cellcolor{gray}96.72/98.05/97.63
\\ \hline

\multirow{2}*{\makecell[c]{\textbf{WRN-28-10} \\ \textbf{(on CIFAR-100)}}} &\textbf{ELOC} \cite{vyas2018out}  
                                           &21.40 &10.48 &95.87 \\
&\cellcolor{gray}\textbf{+SiT/+HiT/+MiT}   &\cellcolor{gray}16.91/13.63/17.73 &\cellcolor{gray}8.75/7.87/9.10 &\cellcolor{gray}97.01/97.47/96.78
\\ \hline

\end{tabular}
} 
\end{center}
\end{table*}

\begin{table*}
\begin{center}
\caption{\textbf{One-class anomaly detection} on CIFAR-10 \cite{krizhevsky2009learning}. Image-level AUROC in $\%$ is given. $\mathbf{class_i}$ indicates the $i$th class. `Extra' indicates utilizing extra data for training (\eg, using pre-trained models on ImageNet\cite{deng2009imagenet}). ``*" indicates results are obtained via self-implemented network. Mixed-geometry embedding in the +MiT model incorporates both the spherical embedding and hyperbolic embedding. RotNet$^{*}$, +SiT, +HiT and +MiT adopts WRN-28-10 \cite{zagoruyko2016wide} as feature extractor while CSI, +SiT, +HiT and +MiT utilizes ResNet-18 \cite{he2016deep}. The curvatures of RotNet$^{*}$+SiT, +HiT and MiT are set to 1.0, -0.005 and (1.0, -0.005). The curvatures of CSI+SiT, +HiT and MiT are given as 1.0, -0.01 and (1.0, -0.01).} \label{table:cifar10}

\resizebox{1.0\textwidth}{!}{
\begin{tabular}{lc|ccccccccccc}  \\ 
\hline
\textbf{Method} &\textbf{Extra} &$\mathbf{class_0}$ &$\mathbf{class_1}$ &$\mathbf{class_2}$ &$\mathbf{class_3}$ &$\mathbf{class_4}$ &$\mathbf{class_5}$ &$\mathbf{class_6}$ &$\mathbf{class_7}$ &$\mathbf{class_8}$ &$\mathbf{class_9}$ &\textbf{Avg} \\ \hline

$\textbf{RotNet}^{*}$ \cite{gidaris2018unsupervised} &\XSolidBrush &72.70 &94.25 &76.38 &69.26 &79.30 &80.97 &76.18 &92.78 &90.62 &88.67 &82.11 \\
\rowcolor{gray} $\textbf{RotNet}^{*}\textbf{+SiT}$  &\XSolidBrush &73.77 &95.21 &78.39 &70.46 &80.92 &81.75 &77.80 &93.25 &91.27 &89.40 &83.22 \\
\rowcolor{gray} $\textbf{RotNet}^{*}\textbf{+HiT}$  &\XSolidBrush &72.68 &94.54 &78.31 &69.62 &80.80 &81.63 &77.05 &93.06 &91.06 &88.90 &82.77 \\
\rowcolor{gray} $\textbf{RotNet}^{*}\textbf{+MiT}$  &\XSolidBrush &73.60 &95.18 &79.82 &70.80 &81.76 &82.57 &78.52 &93.52 &91.16 &89.46 &\textbf{83.64} \\
\hline

$\textbf{CSI}$ \cite{tack2020csi} &\XSolidBrush &89.9  &99.1 &93.1 &86.4 &93.9 &93.2 &95.1 &98.7 &97.9 &95.5 &94.3 \\
\rowcolor{gray} \textbf{CSI+SiT}  &\XSolidBrush &89.26 &99.17 &94.16 &87.76 &94.53 &93.47 &95.44 &98.85 &97.91 &96.03 &94.66 \\
\rowcolor{gray} \textbf{CSI+HiT}  &\XSolidBrush &89.40 &99.17 &94.22 &88.06 &94.33 &93.48 &95.58 &98.76 &97.85 &95.93 &94.68 \\
\rowcolor{gray} \textbf{CSI+MiT}  &\XSolidBrush &89.46 &99.16 &94.30 &88.04 &94.41 &93.32 &95.58 &98.87 &97.94 &95.99 &\textbf{94.71} \\
\hline

\end{tabular}} 
\end{center}
\end{table*}

In this section, we evaluate our models on four visual anomaly or OOD tasks: (1) multi-class OOD detection (in Tabs.~\ref{table:cifar10_100},~\ref{table:cifar10_100_odin} and~\ref{table:cifar10_100_eloc}), (2) one-class anomaly detection (in Tab.~\ref{table:cifar10}), (3) multi-class anomaly segmentation (in Tab.~\ref{table:street}) and (4) one-class anomaly segmentation (in Tabs.~\ref{table:MVAL} and \ref{table:MVAL_supple}). Tab.~\ref{table:tasks} shows the difference of each task used in our paper. All the tasks are all used to evaluated our models. For simplicity, we use the following abbreviations for our models: \textbf{SiO} (Spherical-in-One), \textbf{SiT} (Spherical-in-Two), \textbf{HiO} (Hyperbolic-in-One), \textbf{HiT} (Hyperbolic-in-Two), \textbf{MiO} (Mixed-in-One) and \textbf{MiT} (Mixed-in-Two). See Tab. \ref{table:GIO} for more details. Metrics including the false positive rate at 95\% true positive rate (FPR at 95\% TPR), the detection error, the area under receiver operating characteristics (AUROC) \cite{davis2006relationship,fawcett2006introduction}, and the area under the precision-recall (AUPR) \cite{manning1999foundations,saito2015precision} are measured. All results are averaged over 5 independent trials.

\subsection{Multi-Class OOD Detection} \label{sec:mcoodd}
The objective of multi-class OOD detection, traditionally termed OOD detection, is to identify whether a sample is from the given dataset with multiple ID classes. The model is trained on the ID dataset only. In this setting, CIFAR-10 and CIFAR-100 \cite{krizhevsky2009learning} are chosen as ID datasets while the cropped TinyImageNet (TINc), the resized TinyImageNet (TINr) \cite{deng2009imagenet}, the cropped LSUN (LSUNc), the resized LSUN (LSUNr) \cite{yu2015lsun} and iSUN \cite{xu2015turkergaze} are OOD datasets. We first adopt the Hendrycks$\&$Gimpel model \cite{hendrycks2016baseline} as the baseline network. Both Dense-BC \cite{huang2017densely} and WEN-28-10 \cite{zagoruyko2016wide} are used as backbones. As shown in Tab. \ref{table:cifar10_100}, we can observe that the +HiO model attains the overall best accuracy. In addition, our models, except the +SiO model, bring the performance gain over the baselines, showing the superiority of curved geometries as embedding spaces. It is also notable that our models are light. For example, the +HiT model, surpasses the baseline by 4.2\% with WRN-28-10 on CIFAR-100, while it only uses an extra 0.02M parameters, \ie, from 146.05M to 146.07M. Moreover, in most cases, performance of the mixed-curvature model, +MiO or +MiT, is in between that of hyperbolic and spherical models. 
Besides Hendrycks$\&$Gimpel model, we also use ODIN \cite{liang2017enhancing} as the baseline where we employ the input pre-processing at the test phase. The results of geometric models which adopt ODIN as the baseline are reported in Tab. \ref{table:cifar10_100_odin}. From Tab. \ref{table:cifar10_100}, we identify +HiO is the model which obtains the best performance. Hence, we choose and test +HiO for ODIN. Except the experiment of WRN-28-10 on CIFAR-100, we see +HiO boosts the performance against the baseline.

In addition to Hendrycks$\&$Gimpel and ODIN models, we also incorporate the proposed geometric classifier into advanced baselines. In this study, we employ the ELOC~\cite{vyas2018out} as the baseline network. As shown in Tab. \ref{table:cifar10_100_eloc}, the +HiT model performs the best over two backbones. Specifically, it surpasses the baseline by 0.77\%/1.60\% in AUROC under Dense-BC/WRN-28-10 backbones. Similar to results shown in Tab. \ref{table:cifar10_100}, the performance of the +MiT model is in between +SiT and +HiT with Dense-BC on CIFAR-100, but it unexpectedly becomes the worst with WRN-28-10.

\subsection{One-Class Anomaly Detection}\label{sec:ocad}
In the one-class anomaly detection setting, only one class is set as the normal class while other classes are used as abnormal classes. The common practice of creating discriminative representations under this setting is modeled as a multi-class classification problem, using the self-supervised learning (SSL) algorithms~\cite{gidaris2018unsupervised,ruff2018deep,golan2018deep,hendrycks2019using,tack2020csi,tack2020csi}. In this task, we evaluate our models on the one-class CIFAR-10 dataset \cite{krizhevsky2009learning}.

Our geometric classifier is built on RotNet~\cite{gidaris2018unsupervised} and CSI~\cite{tack2020csi}. RotNet predicts the rotation angles as supervision for SSL. Following the setting in \cite{gidaris2018unsupervised},  we set the rotation angles to 0$^{\circ}$, 90$^{\circ}$, 180$^{\circ}$ and 270$^{\circ}$. A 4-dimension classifier predicts the angle of rotation is applied to the input image. CSI adopts the contrastive learning scheme, which contrasts the negative samples coming from the data augmentation. The results are shown in Tab. \ref{table:cifar10}. We can observe that either of our models can improve the baselines, showing the embedding in curved spaces indeed benefits the discrimination of data embedding. For example, in RotNet, the method with mixed-curvature geometry, +MiT, attains the best performance improvement, \eg, 1.53\%, and outperforms the +SiT and +HiT models. It verifies that the mixed-curvature geometry indeed benefits from the advantages of both spherical geometry and hyperbolic geometry. In CSI \cite{tack2020csi}, as a strong baseline, our models again bring a performance gain and the +MiT method achieves the best average performance, revealing that our models  generalise and are effective.

\subsection{Multi-Class Anomaly Segmentation}\label{sec:mcas}
In contrast to multi-class OOD detection, which recognizes the OOD samples at the image level, multi-class anomaly segmentation is required to predict anomalous objects at the pixel level.  Following \cite{hendrycks2019scaling,xia2020synthesize}, we evaluate this task using the StreetHazards dataset \cite{hendrycks2019scaling}. The Hendrycks$\&$Gimpel model \cite{hendrycks2016baseline} with Maximum Softmax Probability (MSP) is adopted as the baseline. We report the results in Tab. \ref{table:street}. As suggested in Tab. \ref{table:street}, this task also benefits the most from the mixed-curvature geometry (MSP+MiT), again showing that multiple geometries are essential to learning discriminative embeddings.

\begin{table}
\begin{center}
\caption{\textbf{Multi-class anomaly segmentation} on StreetHazards \cite{hendrycks2019scaling}. FPR (95$\%$ TPR), pixel-level AUROC and AUPR in $\%$ are given. The metric std over 5 runs. The results of AE, Dropout and MSP are provided in \cite{hendrycks2019scaling}.}
\label{table:street}

\resizebox{.48\textwidth}{!}{
\begin{tabular}{l|ccc}  \\ 
\hline
\textbf{Method} &\textbf{FPR (95\% TPR) $\downarrow$} &\textbf{AUROC $\uparrow$} &\textbf{AURP $\uparrow$} \\ \hline

\textbf{AE} \cite{baur2018deep}                   &91.7  &66.1  &2.2  \\
\textbf{Dropout} \cite{gal2016dropout}            &79.4  &69.9  &7.5  \\ 
\textbf{MSP} \cite{hendrycks2016baseline}         &33.7  &87.7  &6.6  \\ 
\textbf{SynthCP} \cite{xia2020synthesize}         &28.4  &88.5  &9.3  \\ 
\textbf{MaxLogit} \cite{hendrycks2019scaling}     &\textbf{26.5}  &\textbf{89.3}  &\textbf{10.6}  \\ 

\hline
\rowcolor{gray} \textbf{MSP} \cite{hendrycks2016baseline}       &33.7  &87.7  &6.6  \\ \hdashline
\rowcolor{gray} \textbf{MSP+SiT}        &29.9  &86.4  &4.6  \\
\rowcolor{gray} \textbf{MSP+HiT}        &27.7  &88.9  &6.3  \\
\rowcolor{gray} \textbf{MSP+MiT}        &\textbf{27.1} &\textbf{89.5} &\textbf{8.6}  \\
\hline
\end{tabular}} 
\end{center}
\end{table}

\subsection{One-Class Anomaly Segmentation}\label{sec:ocas}
One-class anomaly segmentation, also known as anomaly localization, aims to identify whether the input pixel is an anomalous pixel or not \cite{bergmann2019mvtec,liu2020towards}. In contrast to the multi-class anomaly segmentation, the training samples in one-class anomaly segmentation are drawn from only one class of the dataset.

\begin{table*}
\begin{center}
\caption{\textbf{One-class anomaly segmentation (anomaly localization)} on MVTecAD \cite{bergmann2019mvtec}. Pixel-level AUROC in $\%$ is given. `Extra' indicates utilizing extra data for training (\eg, using pre-trained models on ImageNet\cite{deng2009imagenet}). `Cpr' indicates using the comparison with the training data for anomaly score computation during evaluation. The results of AnoGAN/AE and AVID are provided in \cite{Salehi_2021_CVPR} and \cite{venkataramanan2020attention}, respectively. Mixed-geometry embedding in the +MiO/+MiT model includes spherical embedding and hyperbolic embedding. The curvatures of +SiO, +HiO, +MiO, +SiT, +HiT and +MiT are set to 1.0, -0.01, (1.0, -0.01), 1.0, -1.0 and (1.0, -1.0). The model sizes of Patch-SVDD$^{\dagger}$, +SiO, +HiO, +MiO, +SiT, +HiT and +MiT are 1.72M, 1.72M, 1.72M, 1.82M, 1.82M, 1.83M and 1.93M.} \label{table:MVAL}

\resizebox{1.00\textwidth}{!}{
\begin{tabular}{lcc|cccccccccccccccc}  \\ 
\hline
\textbf{Method} &\textbf{Extra} &\textbf{Cpr} &\textbf{Bottle} &\textbf{Hazelnut} &\textbf{Capsule} &\textbf{Metal Nut} &\textbf{Leather} 
&\textbf{Pill} &\textbf{Wood} &\textbf{Carpet} &\textbf{Tile} &\textbf{Grid} &\textbf{Cable} &\textbf{Transistor} &\textbf{Toothbrush} &\textbf{Screw} &\textbf{Zipper} &\textbf{Avg} \\ \hline

\textbf{AnoGAN} \cite{schlegl2017unsupervised}            &\XSolidBrush  &\XSolidBrush
&86  &87  &84  &76  &64  &87  &62  &54  &50  &58  &78  &80  &90  &80  &78  &74 \\

\textbf{AVID} \cite{sabokrou2018avid}                     &\XSolidBrush  &\XSolidBrush
&-  &-  &-  &-  &-  &-  &-  &-  &-  &-  &-  &-  &-  &-  &-  &77 \\

\textbf{AE-L2}\cite{bergmann2018improving}                &\XSolidBrush  &\XSolidBrush
&86  &95  &88  &86  &75  &85  &73  &59  &51  &90  &86  &86  &93  &96  &77  &82 \\ 

\textbf{AE-SSIM}\cite{bergmann2018improving}              &\XSolidBrush  &\XSolidBrush
&93  &97  &94  &89  &78  &91  &73  &87  &59  &94  &82  &90  &92  &96  &88  &87 \\

\textbf{VAE-VE} \cite{liu2020towards}                     &\XSolidBrush  &\XSolidBrush
&87  &98  &74  &94  &95  &83  &77  &78  &80  &73  &90  &93  &94  &97  &78  &86 \\

\textbf{VAE-grad} \cite{dehaene2020iterative}             &\XSolidBrush  &\XSolidBrush 
&92.2  &97.6  &91.7  &90.7  &92.5  &93  &83.8  &73.5  &65.4  &96.1  &91.0  &91.9  &98.5  &94.5  &86.9  &89.3 \\ 

\hline
$\textbf{Patch-SVDD}^{\dagger}$ \cite{yi2020patch}   &\XSolidBrush &\XSolidBrush
&62.78 &71.03 &80.03 &68.28	&90.03 &67.15 &65.45 &82.60 &89.28	&82.55	&67.60	&50.35	&88.20 &85.95	&67.23	&74.57 \\ \hdashline

\rowcolor{gray} $\textbf{Patch-SVDD}^{\dagger}$\textbf{+SiO}          &\XSolidBrush &\XSolidBrush
&62.80	&73.53 &80.50	&68.40	&89.70	&70.13	&65.50	&74.10	&85.90	&82.10	&68.35	&51.10	&86.57	&50.00	&68.35	&71.80 \\

\rowcolor{gray} $\textbf{Patch-SVDD}^{\dagger}$\textbf{+HiO}          &\XSolidBrush &\XSolidBrush
&68.27	&76.43	&89.00	&50.00	&93.65	&73.20	&79.90	&86.40	&86.65	&78.50	&87.45	&71.50	&87.03	&78.65	&90.60	&79.82 \\

\rowcolor{gray} $\textbf{Patch-SVDD}^{\dagger}$\textbf{+MiO}          &\XSolidBrush &\XSolidBrush
&68.50	&79.73	&92.03	&65.28	&94.80	&74.83	&79.00	&86.20	&70.20	&66.17	&61.93	&64.47	&87.43	&71.10	&88.08	&76.65 \\ \hline

\rowcolor{gray} $\textbf{Patch-SVDD}^{\dagger}$\textbf{+SiT}          &\XSolidBrush &\XSolidBrush
&95.43	&94.60	&90.80	&96.07	&96.30	&92.20	&82.93	&89.20	&92.55	&89.13	&93.67	&92.40	&93.30	&83.40	&91.68	&91.58 \\

\rowcolor{gray} $\textbf{Patch-SVDD}^{\dagger}$\textbf{+HiT}          &\XSolidBrush &\XSolidBrush
&95.80  &93.60  &92.87  &88.40  &97.03  &92.90  &81.10  &83.13  &87.03  &92.03  &95.20  &95.27  &94.57  &81.63  &94.20  &90.98 \\ 

\rowcolor{gray} $\textbf{Patch-SVDD}^{\dagger}$\textbf{+MiT}          &\XSolidBrush &\XSolidBrush
&96.33	&96.05	&92.77	&96.77	&96.63	&93.50	&82.30	&84.30	&93.63	&93.70	&94.93	&94.40	&95.03	&77.33	&94.13	&\textbf{92.12} \\
\hline

\end{tabular}} 
\end{center}
\end{table*}

\begin{table*}
\begin{center}
\caption{\textbf{One-class anomaly segmentation (anomaly localization)} on MVTecAD \cite{bergmann2019mvtec}. Pixel-level AUROC in $\%$ is given. `Extra' indicates utilizing extra-data for training (\eg, using pre-trained models on ImageNet\cite{deng2009imagenet}). `Cpr' indicates using the comparison with the training data for anomaly score computation during evaluation. The results of LSA are provided in \cite{venkataramanan2020attention}. Mixed-geometry embedding in the +MiT model includes spherical embedding and hyperbolic embedding. The curvatures of +SiT, +HiT and +MiT are set to 1.0, -1.0 and (1.0, -1.0).} \label{table:MVAL_supple}

\resizebox{1.00\textwidth}{!}{
\begin{tabular}{lcc|cccccccccccccccc}  \\ 
\hline
\textbf{Method}  &\textbf{Extra} &\textbf{Cpr} &\textbf{Bottle} &\textbf{Hazelnut} &\textbf{Capsule} &\textbf{Metal Nut} &\textbf{Leather} 
&\textbf{Pill} &\textbf{Wood} &\textbf{Carpet} &\textbf{Tile} &\textbf{Grid} &\textbf{Cable} &\textbf{Transistor} &\textbf{Toothbrush} &\textbf{Screw} &\textbf{Zipper} &\textbf{Avg} \\ \hline

\textbf{LSA}\cite{Abati_2019_CVPR}                        &\Checkmark    &\XSolidBrush
&-  &-  &-  &-  &-  &-  &-  &-  &-  &-  &-  &-  &-  &-  &-  &81 \\ 

\textbf{CAVGA-Rw} \cite{venkataramanan2020attention}      &\Checkmark    &\XSolidBrush  
&-  &-  &-  &-  &-  &-  &-  &-  &-  &-  &-  &-  &-  &-  &-  &90 \\

\textbf{MR-KD} \cite{Salehi_2021_CVPR}                    &\Checkmark                     &\XSolidBrush
&96.32  &94.62  &95.86  &86.38  &98.05  &89.63  &84.80  &95.64  &82.77  &91.78  &82.40  &76.45  &96.12  &95.96  &93.90 &90.71 \\

\textbf{Glancing-at-Patch} \cite{Wang_2021_CVPR}          &\Checkmark    &\XSolidBrush  
&93  &84  &90  &91  &90  &93  &81  &96  &80  &78  &94  &100  &96  &96  &99 &91 \\

\textbf{PANDA-OE} \cite{Reiss_2021_CVPR}                  &\Checkmark   &\XSolidBrush  
&-  &-  &-  &-  &-  &-  &-  &-  &-  &-  &-  &-  &-  &-  &-  &96.2 \\

\textbf{CutPaste} \cite{li2021cutpaste}                    &\XSolidBrush &\Checkmark  
&97.6  &97.3  &97.4  &93.1  &99.5  &95.7  &95.5  &98.3  &90.5  &97.5  &90.0  &93.0  &98.1  &96.7  &99.3  &96.0 \\ \hline

                $\textbf{Patch-SVDD}$ \cite{yi2020patch}  &\XSolidBrush &\Checkmark
&98.1  &97.5  &95.8  &98.0  &97.4  &95.1  &90.8  &92.6  &91.4  &96.2  &96.8  &97.0  &98.1  &95.7  &95.1  &95.7 \\ \hdashline

\rowcolor{gray} $\textbf{Patch-SVDD}$\textbf{+SiT}        &\XSolidBrush &\Checkmark
&98.10  &98.93  &97.00  &98.00  &97.20  &94.83  &92.30  &91.20  &96.30  &97.43  &96.30  &97.83  &96.40  &97.53  &98.45  &96.52 \\

\rowcolor{gray} $\textbf{Patch-SVDD}$\textbf{+HiT}        &\XSolidBrush &\Checkmark
&98.23  &96.90  &96.83  &97.60  &97.20  &94.20  &92.73  &94.13  &96.50  &97.65  &95.75  &97.60  &95.95  &97.37  &98.37  &96.47 \\

\rowcolor{gray} $\textbf{Patch-SVDD}$\textbf{+MiT}        &\XSolidBrush &\Checkmark
&98.23  &99.00  &96.33  &98.80  &97.20  &96.03  &92.43  &91.70  &96.60  &97.50  &96.80  &97.63  &96.65  &97.73  &98.45  &\textbf{96.74} \\

\hline
\end{tabular}} 
\end{center}
\end{table*}

We verify the effectiveness of our models on the MVTecAD dataset \cite{bergmann2019mvtec}, and adopt the SOTA model Patch-SVDD \cite{yi2020patch} as a baseline without using extra data. A possible limitation of Patch-SVDD is that the computation of the anomaly scores $\text{AS}$ for inference depends to a great extent on the comparison with training samples. To simplify the evaluation process, we calculate the anomaly score directly from its normalized classifier's value without utilizing any training data (denoted by Patch-SVDD$^{\dagger}$). We then plug our geometric models on top of Patch-SVDD$^{\dagger}$. The results are reported in Tab. \ref{table:MVAL}. As suggested by Tab. \ref{table:MVAL}, all geometric models, except the +SiO model, boost the performance of the baseline, and the mixed-curvature geometric model (\ie, Patch-SVDD$^{\dagger}${+MiT}) performs the best. It gains 17.55\% improvement. In this task, the dual-branch architecture (\eg, +SiT/+HiT/+MiT) consistently outperforms the single-branch model (\ie, +SiO/+HiO/+MiO). Along with a considerable improvement, our proposal is also cheap. For example, the Patch-SVDD$^{\dagger}${+SiT} improves the baseline by a margin of 16.41\%, while it only brings extra 0.1M parameters, \ie,  1.82M vs. 1.72M, again showing the benefits from curved geometric embeddings. 

The idea of Patch-SVDD$^{\dagger}$ is to evaluate our method on a toy example to illustrate the advantage of curved geometries. However, after we remove the comparison process with training images from Patch-SVDD, we find its identification performance significantly drops. To show the full potential of our design in conjunction with the original Patch-SVDD, we employ our geometric model over the original Patch-SVDD where training images are considered for calculating the anomaly score. As shown in Tab. \ref{table:MVAL_supple}, the accuracy of Patch-SVDD on MVTecAD is boosted from 95.7\% to 96.5\%/96.5\%/96.7\% once combined with SiT/HiT/MiT. We follow the anomaly score computation of the original Patch-SVDD except for replacing patch's embedding with the geometric score $z_{EG}$.

\subsection{Analysis}
In this part, we aim to provide studies to analyze the superiority of our design.

\begin{figure*}
\begin{center}
	\includegraphics[width=18cm]{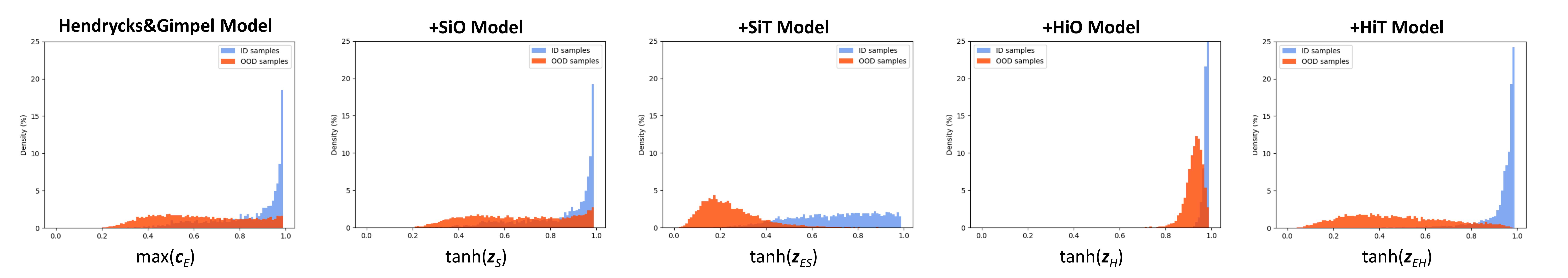}
 	\caption{\textbf{Visualization: multi-class OOD detection.} Density distributions of $\mathrm{max}(\textbf{c}_E)$, $\mathrm{tanh}(z_{S})$, $\mathrm{tanh}(z_{ES})$, $\mathrm{tanh}(z_{H})$ and $\mathrm{tanh}(z_{EH})$ of Hendrycks\&Gimpel, +SiO, +SiT, +HiO and +HiT models with Dense-BC backbone on CIFAR10$\rightarrow$TINc are provided. The obtained AUROC for these five models are 94.8, 94.7, 98.1, 98.7 and 96.0. More corresponding results can be referred to in Tab. \ref{table:cifar10_100}.}\label{fig:visual-1}
\end{center}
\end{figure*}

\begin{figure}
\begin{center}
	\includegraphics[width=8.6cm]{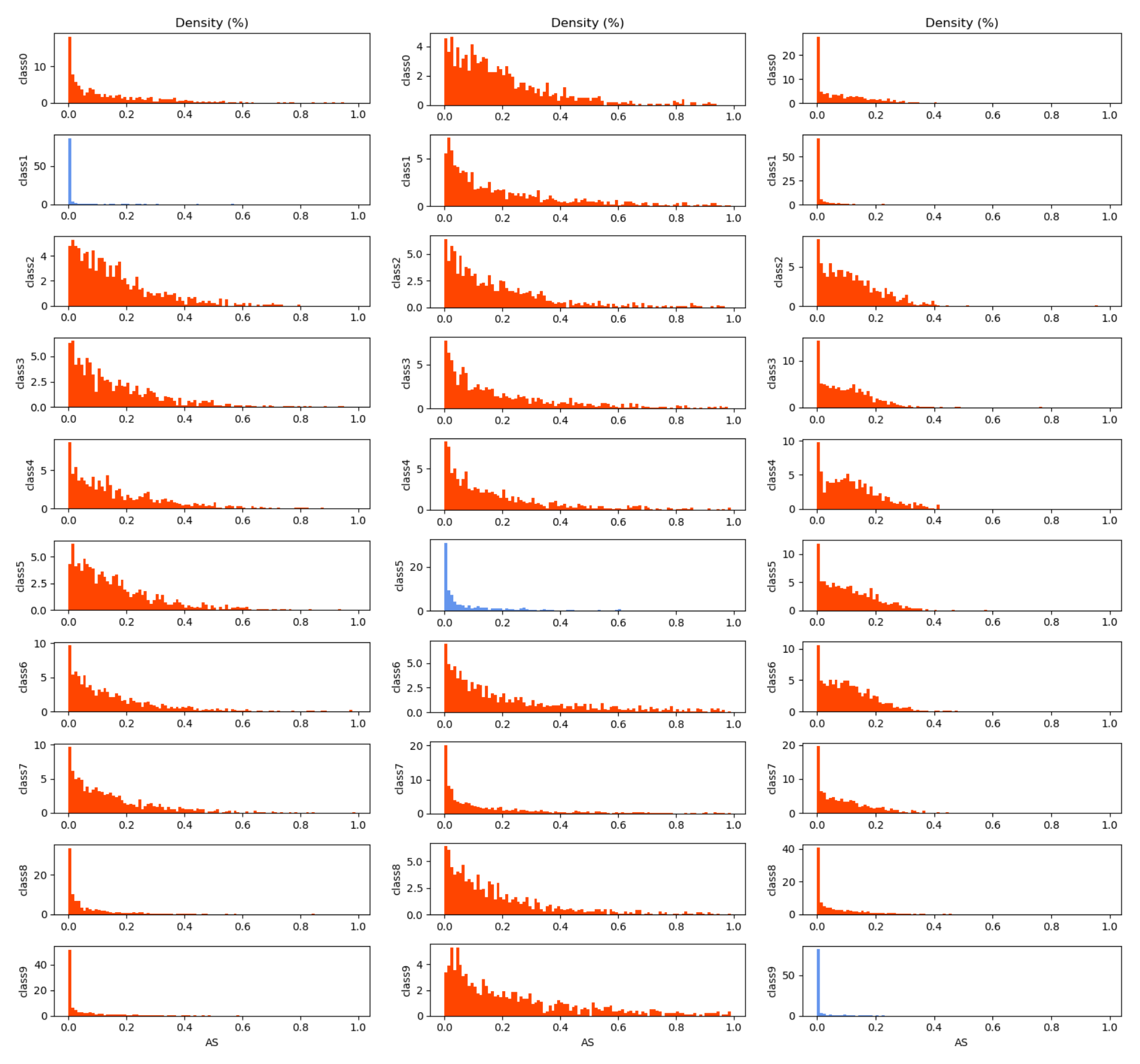}
 	\caption{\textbf{Visualization: one-class anomaly detection.} Density distribution of anomaly score $\text{AS}(z_{EH})$ of RotNet$^{*}$+HiT on CIFAR-10 \cite{krizhevsky2009learning} is visualized.  We choose cases where $\mathbf{class_1}$, $\mathbf{class_5}$ or $\mathbf{class_9}$ is taken as the normal class. The distribution of one case is shown in one column. More corresponding results can be referred to in Tab. \ref{table:cifar10}}\label{fig:visual-2}
\end{center}
\end{figure}

\noindent \textbf{Performance.} We learn from empirical observations in \textsection~\ref{sec:mcoodd},~\ref{sec:ocad},~\ref{sec:mcas} and~\ref{sec:ocas} that the curved spaces are able to consistently provide reliable information for anomaly or OOD recognition. In most cases, our curvature-aware geometric networks clearly outperform Euclidean networks. 
One possible explanation is that the geometric representations benefit particular problems, \eg, hyperbolic geometry is good at encoding hierarchical structures inside the data. We hypothesize, the datasets we test include the hierarchical information to some extent. For instance, CIFAR dataset includes 10 super-classes and 10 sub-classes under each super-class. The 15 classes of MVTecAD can be categorised into two main super-categories, `object' and `texture'. Hence, the hyperbolic space takes effects.

\noindent \textbf{Mixed geometry.} Another interesting fact has been observed is that the mixed-curvature geometry beats its component single-curvature geometries in several cases: RotNet$^{*}$/CSI+MiT of one-class anomaly detection, MSP+MiT of multi-class anomaly segmentation and Patch-SVDD$^{\dagger}$+MiT of one-class anomaly segmentation in Tab. \ref{table:cifar10}, \ref{table:street} and \ref{table:MVAL}, respectively. In some cases, the mixed-curvature geometry has the balanced performance. For example, in the task of multi-class OOD detection, there exists a significant performance gap between hyperbolic and spherical geometries, as evidenced by +SiO \vs +HiO. Thus, the mixed space +MiO might be expected to have an averaged performance.

\noindent \textbf{Interactions among geometries.} The GiT model requires meanwhile learning two embeddings (\eg, a Euclidean $\textbf{e}_{E}$, and a hyperbolic $\textbf{e}_{H}$ or spherical embedding $\textbf{e}_{S}$). 
We observe that it happens the interactions between different geometric components. For instance, in multi-class OOD detection (See Tab. \ref{table:cifar10_100} in \textsection~\ref{sec:mcoodd}), $\textbf{e}_{E}$ could enrich $\textbf{e}_{S}$ (+SiO \vs +SiT), however, for $\textbf{e}_{H}$, it has less or even negative impact (+HiO \vs +HiT). To further understand the influence, we analyzed the experiments of WRN-28-10 on CIFAR-10 where we separately test $\textbf{e}_{E}$ and $\textbf{e}_{H}$ (or $\textbf{e}_{S}$) in +HiT (or +SiT). Results in Tab. \ref{table:ablation3} suggests the  aforementioned point (+HiO \vs $\textbf{e}_{H}$ in +HiT and +SiO \vs $\textbf{e}_{S}$ in +SiT).  

\begin{table}
  \centering
  \caption{\textbf{Ablation study: multi-class OOD detection.} Image-level AUROC in $\%$ is given. Different cases are evaluated on CIFAR-10 \cite{krizhevsky2009learning} to verify the interactions between hyperbolic or spherical and Euclidean spaces.} \label{table:ablation3}
  
  \resizebox{0.3\textwidth}{!}{
  \begin{tabular}{c|ccc}
    \hline
    \textbf{+HiO} &\textbf{+HiT}  &\textbf{$\textbf{e}_{E}$ in +HiT} &\textbf{$\textbf{e}_{H}$ in +HiT} \\ \hline
     \rowcolor{gray} 97.1 &94.1 &92.1 &94.5  \\ \hline  
    
    \textbf{+SiO} &\textbf{+SiT}   &\textbf{$\textbf{e}_{E}$ in +SiT} &\textbf{$\textbf{e}_{S}$ in +SiT} \\ \hline
    \rowcolor{gray} 93.4 &97.1  &93.6  &96.8
    \\ \hline
  \end{tabular}}
\end{table}

\noindent \textbf{Curvature.} The curvature $\kappa$ is the only hyper-parameter in the proposed curvature-aware geometric networks. The study of one-class anomaly segmentation of Patch-SVDD$^{\dagger}$+HiT in Tab. \ref{table:ablation1} suggests that $\kappa$ clearly has an impact on the anomaly recognition performance. 

\noindent \textbf{Method Choice.} Our comprehensive empirical study suggests that a single definitive conclusion cannot be made. This is in line with observations made in recent works. For example, in \cite{MixedVAE_ICLR}, the best geometry choice depends on the task. Our study clearly shows that curved geometry is beneficial in capturing the geometry of data, contributing tangibly to identifying anomalies in data.  
Specifically, our empirical study suggests that the preferred model for each task is shown below: Multi-class OOD Detection: +HiO/+HiT; Multi-class Anomaly Segmentation: +MiT; One-class Anomaly Detection: +MiT; One-class Anomaly Segmentation: +MiT. If one model should be chosen in all instances, then we will opt for +MiT as the potential model for the visual anomaly recognition tasks.

\begin{table}
\begin{center}
\caption{\textbf{Ablation study: one-class anomaly segmentation (anomaly localization).} Pixel-level AUROC in $\%$ is given. Different fixed curvatures on MVTecAD \cite{bergmann2019mvtec} are evaluated. For each category, different curvatures brings about different performances. Also, the optimal curvature choice is not consistent among different categories.} \label{table:ablation1}

\resizebox{.48\textwidth}{!}{
\begin{tabular}{l|c|cccc}  
\hline
\textbf{Method} &\textbf{$\kappa$} &\textbf{Hazelnut} &\textbf{Leather} &\textbf{Wood} &\textbf{Toothbrush} \\ \hline

\rowcolor{gray} $\textbf{Patch-SVDD}^{\dagger}$\textbf{+HiT}        &-$1\mathrm{e}{-4}$    &93.77  &96.30  &\textbf{84.67}  &89.90 \\
\rowcolor{gray} $\textbf{Patch-SVDD}^{\dagger}$\textbf{+HiT}        &-$1\mathrm{e}{-2}$    &\textbf{95.40}  &95.77  &83.87  &88.83 \\
\rowcolor{gray} $\textbf{Patch-SVDD}^{\dagger}$\textbf{+HiT}        &-1.0                  &93.60  &\textbf{97.03}  &81.10  &\textbf{94.57} \\
\hline

\end{tabular}} 
\end{center}
\end{table}

\subsection{Visualization}
In this part, we qualitatively study our method in image-level classification task and pixel-level segmentation task, to understand why our method can bring performance gain over the baseline model.

\begin{figure}
\centering
\includegraphics[width=0.9\linewidth]{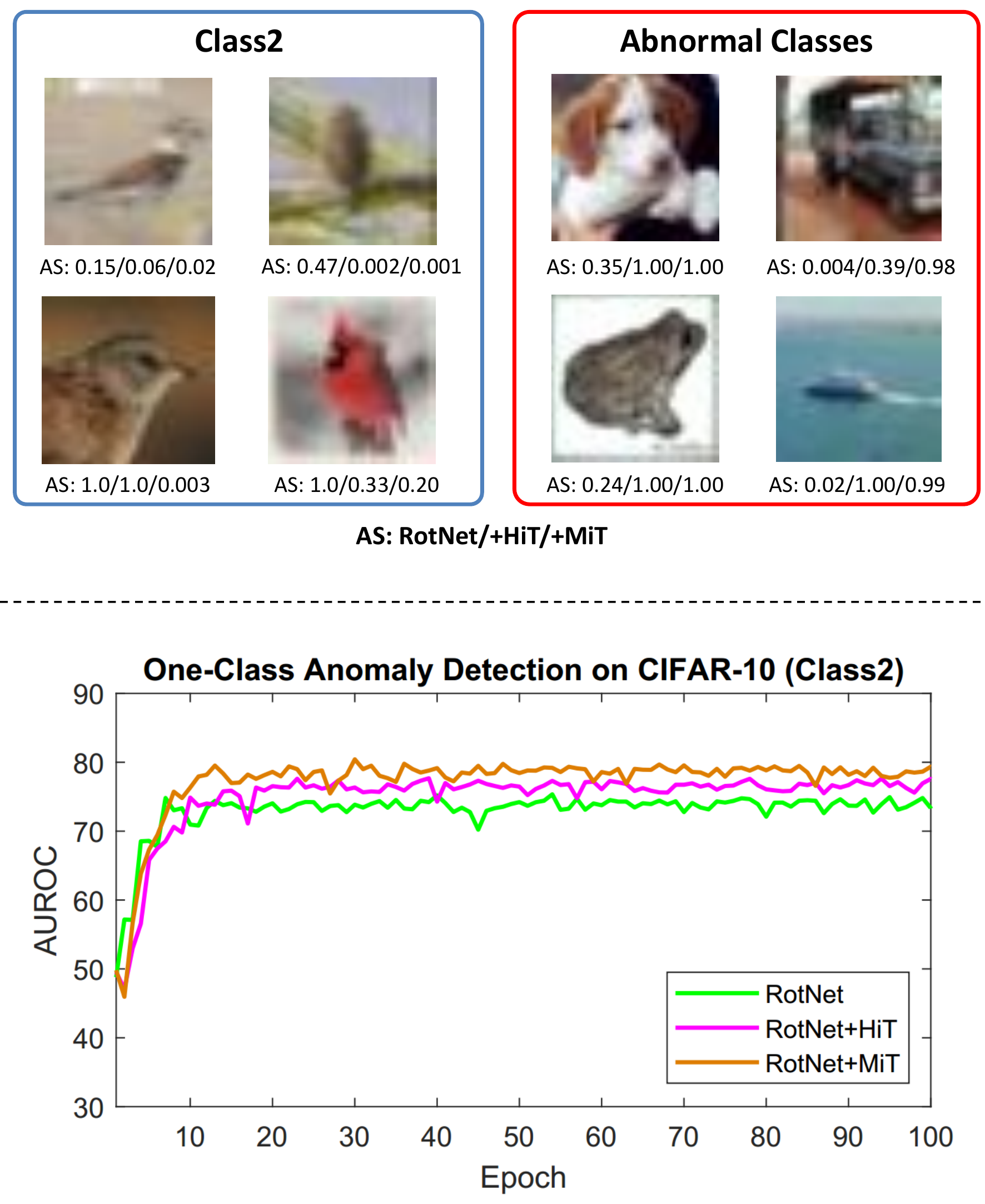}
\caption{\textbf{Visualization: one-class anomaly detection.} Examples with anomaly scores $\text{AS}$ computed by RotNet$^{*}$, +HiT and +MiT are provided (See upper figures). In addition, AUROC curves along the training epoch are plotted in the lower figure.}\label{fig:visual-3}
\end{figure}

\noindent \textbf{Image-Level Classification.} In this work, we particularly study whether the geometric score $z_{G}$ and $z_{EG}$ from the curved embedding spaces can provide more useful information than the confidence vector $\textbf{c}_E$ from the Euclidean space in distinguishing normal (or ID) and abnormal (or OOD) objects. We visualize the distribution of $\mathrm{max}(\textbf{c}_E)$, $\mathrm{tanh}(z_{S})$, $\mathrm{tanh}(z_{ES})$, $\mathrm{tanh}(z_{H})$ and $\mathrm{tanh}(z_{EH})$ on multi-class OOD detection over CIFAR-10 (CIFAR10$\rightarrow$TINc) in Fig. \ref{fig:visual-1}. As shown in Fig. \ref{fig:visual-1}, the curved embedding spaces help to better separate ID and OOD distributions. The distribution of anomaly score $\text{AS}(z_{EH})$ of one-class anomaly detection on one-class CIFAR-10 is visualized in Fig. \ref{fig:visual-2}. The visualization shows $z_{EH}$ provides reliable information for distinguishing the normal one-class and the abnormal classes. We present some examples in Fig. \ref{fig:visual-3} to verify the performance differences among models, where we compare RotNet$^{*}$/+HiT/+MiT on \textit{Class 2} as a normal class. The anomaly scores $\text{AS}$ in the upper figure show different models have different capacities to recognize abnormal classes. Also, the AUROC curve suggests substantial improvements led by the curved geometries.

\begin{figure}
\begin{center}
	\includegraphics[width=8.6cm]{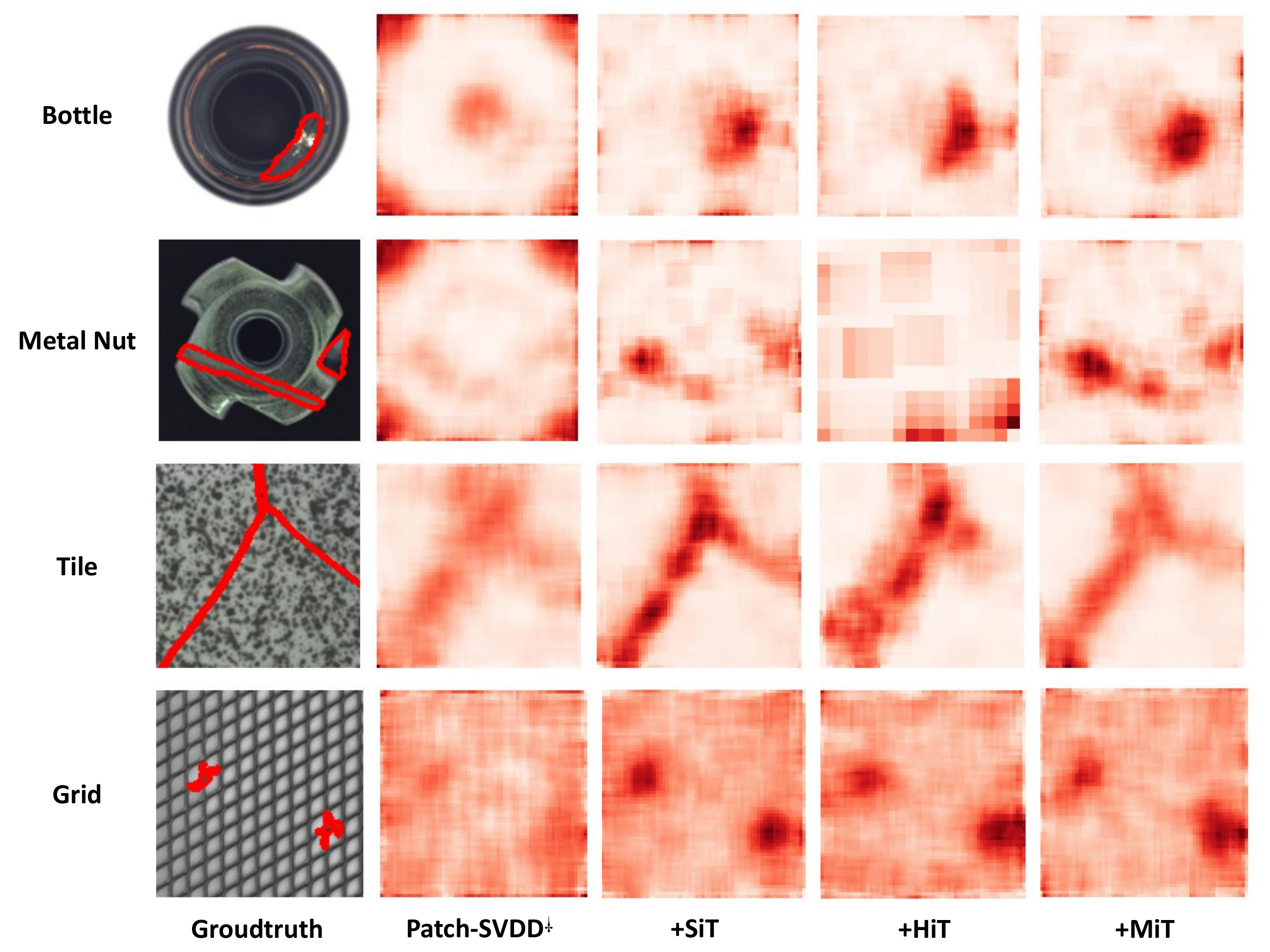}
 	\caption{\textbf{Visualization: one-class anomaly segmentation (anomaly localization).} The heatmaps of anomaly score $\text{AS}(\textbf{c}_E)$, $\text{AS}(z_{ES})$, $\text{AS}(z_{EH})$ and $\text{AS}(z_{EM})$ of Patch-SVDD$^{\dagger}$, +SiT, +HiT and +MiT on MVTecAD \cite{bergmann2019mvtec} are visualized. We provide examples from `bottle', `metal nut', `tile' and `grid'. More corresponding results can be found in Tab. \ref{table:MVAL}.
 	 }\label{fig:visual-4}
\end{center}
\end{figure}

\noindent \textbf{Pixel-Level Segmentation.} From experiments, we find that besides image-level tasks, curved spaces as embedding spaces do help in pixel-level anomaly tasks. In Fig. \ref{fig:visual-4}, we show examples in which we visualize the anomaly score $\text{AS}$ of Patch-SVDD$^{\dagger}$, +SiT, +HiT and +MiT on MVTecAD \cite{bergmann2019mvtec}. As shown in Fig. \ref{fig:visual-4}, $z_{EG}$ from $\mathcal{M}_{G}$ outperforms $\textbf{c}_E$ from $\mathcal{M}_{E}$ in identifying anomalous pixels.

%% file: Conclusion.tex
\section{Conclusion}
In this paper, we study the potential use, and benefit, of employing curved spaces for the purpose of visual anomaly or OOD recognition tasks. Our idea is inspired by the observation that curved embedding spaces help better represent `unknown' data in low-shot problems. Our work proposes two novel geometric networks, geometric-in-one, and geometric-in-two, for the visual anomaly data analysis. In each geometric model, we fully study the potential of different geometry constraints. To the best of our knowledge, our curvature-aware geometric networks are the first attempt to employ curved geometries in visual anomaly or OOD recognition. Based on extensive experiments, we confirm that more distinct representations between normal (or ID) and anomalous (or OOD) samples can be learned using curved spaces, clearly showing the benefits of the curved spaces. 

Though empirical results show the superiority of our proposed method, the theoretical foundation for our findings needs to be further developed to better explain the observations. 
Additionally, the use of the curved embeddings in the generative-based models with `encoder-decoder' structures could be explored in the future work. We hope that this work can inspire researchers to explore curved geometries further in other domains.